%% file: iclr2021_conference.tex
\documentclass{article} 
\usepackage[usenames,dvipsnames]{xcolor}
\usepackage{iclr2021_conference,times}

\usepackage[utf8]{inputenc} 
\usepackage[T1]{fontenc}    
\usepackage{hyperref}       
\usepackage{url}            
\usepackage{booktabs}       
\usepackage{amsfonts}       
\usepackage{nicefrac}       
\usepackage{microtype}      

\input{math_commands.tex}

\usepackage{amsmath}
\usepackage{graphicx}
\usepackage{cleveref}
\usepackage{subcaption}
\usepackage{enumitem}       
\usepackage{commath}
\usepackage{textcomp}
\usepackage{gensymb}       
\usepackage{wrapfig}
\usepackage[font=small,labelfont=bf]{caption}
\usepackage{siunitx}  
\usepackage{floatrow}
\usepackage{refcount}

\usepackage[linesnumbered,ruled,noend]{algorithm2e}

\definecolor{lightgray}{rgb}{.8,.8,.8}
\definecolor{darkgreen}{rgb}{0,.75,0}

\usepackage[colorinlistoftodos, textsize=tiny, textwidth=34mm]{todonotes}
\definecolor{wbcolor}{rgb}{.5, 1, .5}

\newcommand{\Dfm}[0]{\mathcal{D}^{f,m}}
\newcommand{\phii}[0]{\phi^{f,m}_{inter}(x)}
\newcommand{\phif}[0]{\phi^{f,m}_{final}(x)}

\usepackage{tikz}

\usepackage[acronym,smallcaps,nowarn,section,nogroupskip,nonumberlist]{glossaries}
\setkeys{glslink}{hyper=false}
\newacronym{RL}{rl}{Reinforcement Learning}
\newacronym{SL}{sl}{Supervised Learning}
\newacronym{VIB}{vib}{Variational Information Bottleneck}
\newacronym{IB}{ib}{Information Bottleneck}
\newacronym{IBAC}{ibac}{Information Bottleneck Actor Critic}
\newacronym{ITER}{iter}{Iterated Relearning}
\newacronym{RSM}{rsm}{representational similarity matrix}
\newacronym{Alt}{Alt}{Alt}
\glsunset{Alt}
\newacronym{SNI}{sni}{Selective Noise Injection}
\newacronym{SGD}{sgd}{Stochastic Gradient Descent}
\newacronym{DQN}{dqn}{Deep Q-Network}
\newacronym{PPO}{ppo}{Proximal Policy Optimization}
\newacronym{SVD}{svd}{Singular Value Decomposition}
\newacronym[firstplural=Markov decision processes]{MDP}{mdp}{Markov decision process}
\newacronym[firstplural=partially observable Markov decision processes]{POMDP}{pomdp}{partially observable Markov decision process}

\newacronym{IBAC-SNI}{IBAC-SNI}{}
\glsunset{IBAC-SNI}
\newacronym{PG}{pg}{...}
\glsunset{PG}
\newacronym{AC}{ac}{...}
\glsunset{AC}
\newacronym{KL}{kl}{...}
\glsunset{KL}

\def\packenumtopsep{0.75\topsep}
\newenvironment{packenum}{\vspace{-\packenumtopsep}\begin{enumerate}\setlength\itemsep{-0.75mm}}{\end{enumerate}\vspace{-\packenumtopsep}}

\usepackage{amsthm}

\title{Transient Non-stationarity and Generalisation in Deep Reinforcement Learning}

\author{%
Maximilian Igl \thanks{University of Oxford. Corresponding author: Maximilian Igl (\href{mailto:maximilian.igl@gmail.com}{maximilian.igl@gmail.com})}\\
\And
Gregory Farquhar \footnotemark[\value{footnote}] $~^,$\thanks{Now at DeepMind, London}\\
\And
\setcounter{footnote}{1}
Jelena Luketina \footnotemark[\value{footnote}]\\
\And
\setcounter{footnote}{2}
Wendelin B\"ohmer \thanks{Delft University of Technology}\\
\And
\setcounter{footnote}{1}
\centering
Shimon Whiteson \footnotemark[\value{footnote}]
}

\iclrfinalcopy 
\begin{document}

\maketitle

\begin{abstract}
   Non-stationarity can arise in \gls{RL} even in stationary environments.
   For example, most \gls{RL} algorithms collect new data throughout training, using a non-stationary behaviour
   policy.
   Due to the transience of this non-stationarity, it is often not explicitly addressed in deep
   \gls{RL} and a single neural network is continually updated.
   However, we find evidence that neural networks exhibit a memory effect where these transient non-stationarities can
   permanently impact the latent representation and adversely affect generalisation performance.
   Consequently, to improve generalisation of deep \gls{RL} agents, we propose \gls{ITER}.
   \gls{ITER} augments standard \gls{RL} training by repeated knowledge transfer of the current
   policy into a freshly initialised network, which thereby experiences less non-stationarity during
   training.
   Experimentally, we show that \gls{ITER} improves performance on the challenging generalisation
   benchmarks \emph{ProcGen} and \emph{Multiroom}.
\end{abstract}

\input{core/1_intro}

\input{core/2_background}
\input{core/3_nonstat}

\input{core/4_iter}
\input{core/5_experiments}

\input{core/6_relatedWork}
\input{core/7_conclusions}

\newpage

\bibliography{refs}
\bibliographystyle{iclr2021_conference}

\appendix
\newpage
\input{core/9_appendix}

\end{document}

%% file: math_commands.tex
\usepackage{amsmath,amsfonts,bm}

\def\eqref#1{equation~\ref{#1}}

\def\1{\bm{1}}

\DeclareMathAlphabet{\mathsfit}{\encodingdefault}{\sfdefault}{m}{sl}
\SetMathAlphabet{\mathsfit}{bold}{\encodingdefault}{\sfdefault}{bx}{n}

\newcommand{\R}{\mathbb{R}}



%% file: core/1_intro.tex
\section{Introduction}
    \vspace{-0.5em}

In \gls{RL}, as an agent explores more of its environment and updates its
policy and value function, the data distribution it uses for training changes.
In deep \gls{RL}, this non-stationarity is often not addressed explicitly.
Typically, a single neural network model is initialised and continually 
updated during training.
Conventional wisdom about catastrophic forgetting 
\citep{kemker2018measuring} implies 
that old updates from a different data-distribution will simply be forgotten.
However, we provide evidence for an alternative hypothesis:  
networks exhibit a memory effect in their learned representations which can harm generalisation permanently if the
data-distribution changed over the course of training.


To build intuition, we first study this phenomenon in a supervised setting on the CIFAR-10
dataset. 
We artificially introduce transient non-stationarity into the training data and investigate
how this affects the asymptotic performance under the final, stationary data in the later epochs of training.
Interestingly, we find that while asymptotic training performance is nearly unaffected, 
test performance degrades considerably, even after the data-distribution has converged.
In other words, we find that latent representations in deep networks learned under certain types of
non-stationary data can be inadequate for good generalisation and might not be improved by
later training on stationary data.

Such transient non-stationarity is typical in \gls{RL}.
Consequently, we argue that this observed degradation of generalisation might contribute to the inferior
generalisation properties recently attributed to many \gls{RL} agents evaluated 
on held out
test environments \citep{zhang2018dissection,zhang2018study,zhao2019investigating}. 
Furthermore, in contrast to \gls{SL}, simply re-training the agent from scratch once the
data-distribution has changed is infeasible in \gls{RL} as current state of the art algorithms require
data close to the on-policy distribution, even for off-policy algorithms like Q-learning \citep{fedus2020revisiting}.

To improve generalisation of \gls{RL} agents despite this restriction, we propose \glsentryfull{ITER}.
In this paradigm for deep \gls{RL} training, the agent's policy and value are 
periodically distilled into a 
freshly initialised student, which subsequently replaces the teacher for 
further optimisation.
While this occasional distillation step simply aims to re-learn and replace the 
current policy and value outputs for the training data, it
allows the student to learn a better latent representation with improved performance
for unseen inputs because it eliminates non-stationarity during 
distillation.
We propose a practical implementation of \gls{ITER} which performs the distillation in parallel to
the training process without requiring additional training data.
While this introduces a small amount of non-stationarity into the distillation step, it
greatly improves sample efficiency without noticeably impacting performance.

Experimentally, we evaluate \gls{ITER} on the \emph{Multiroom} environment, 
as well as several environments from the recently proposed 
\emph{ProcGen} benchmark and find that it improves generalisation. 
This provides further support to our hypothesis and indicates that the 
non-stationarity inherent to many \gls{RL} algorithms, even when training on
stationary environments, should not be ignored when aiming to learn robust agents. 
Lastly, to further support this claim and provide more insight into possible causes of the
discovered effect, we perform additional ablation studies on the CIFAR-10 dataset.

%% file: core/2_background.tex
\section{Background}
\vspace{-0.5em}
\label{sec:background}
We describe an \gls{RL} problem as a \gls{MDP} $(\mathcal{S}, \mathcal{A}, T, 
r, p_0, \gamma)$
\citep{puterman2014markov} with actions $a\!\in\!\mathcal{A}$, 
states $s\!\in\!\mathcal{S}$, initial state $s_0\!\sim\!p_0$,
transition dynamics $s'\!\sim\!T(s,a)$, 
reward function $r(s,a)\!\in\!\R$ and discount factor $\gamma$.  
The unnormalised discounted state distribution induced by a policy $\pi$ is defined as 
$d_\pi(s)=\sum_{t=0}^\infty \gamma^t \text{Pr}\left(S_t\!=\!s|S_0\!\sim\!p_0, 
A_t\!\sim\!\pi(\cdot|S_t), S_{t+1}\!\sim\!T(S_t, A_t)\right)$. 
In \gls{ITER}, we learn a sequence of policies and value functions, which 
we denote with $\pi^{(k)}(a|s)$ and $V^{(k)}(s)$ at the $k$th iteration ($k\in 
\{0, 1, 2, \dots\}$), parameterized by $\theta_k$.

We briefly discuss some forms of non-stationarity which can arise in \gls{RL}, even when the
environment is stationary. 
For simplicity, we focus the exposition on actor-critic methods which use 
samples from interaction with the environment to estimate the policy gradient 
given by
$g=\mathbb{E}[\nabla_\theta \log \pi_\theta(a|s)A^\pi(s,a,s') | s, a, s' \sim
d_\pi(s)\pi(a|s)T(s'|s,a)]$. 
The advantage is often estimated as $A^\pi(s,a,s') = r(s,a) + \gamma V^\pi(s') - V^\pi(s)$. 
Typically, we also use neural networks to approximate the baseline 
$V^\pi_\phi(s)$ and for bootstrapping from the future value $V_\phi^\pi(s')$.
$\phi$ can be learned by minimising $\mathbb{E}[A^\pi(s,a,s')^2]$ by stochastic 
semi-gradient descent, treating $V_\phi^\pi(s')$ as a constant.

There are at least three main types of non-stationarity in deep \gls{RL}.
First, we update the policy $\pi_\theta$, which leads to changes in the state distribution
$d_{\pi_\theta}(s)$.
Early on in training, a random policy $\pi_\theta$ only explores states close to
initial states $s_0$. As $\pi_\theta$ improves, new states further from $s_0$ are encountered.
Second, changes to the policy also change the true value function $V^\pi(s)$ 
which $V_\phi^\pi(s)$ is approximating.
Lastly, due to the use of bootstrap targets in temporal difference learning, 
the learned value $V_\phi^\pi(s)$ is not regressed directly towards $V^\pi(s)$.
Instead $V_\phi^\pi$ fits a gradually evolving target sequence even under a 
fixed policy $\pi$, thereby also changing the policy gradient estimator $g$. 

%% file: core/3_nonstat.tex
\section{The Impact of Non-stationarity on Generalisation}
\vspace{-0.5em}
\label{sec:nonstat}

In this section we investigate how asymptotic performance is
affected by changes to the data-distribution during training. 
In particular, we assume an initial, transient phase of non-stationarity, 
followed by an extended phase of training on a stationary data-distribution.
This is similar to the situation in \gls{RL} where the data-distribution is affected by a
policy which converges over time.
We show that this transient non-stationarity has a permanent effect on the
learned representation and negatively impacts generalisation.
As interventions in \gls{RL} training can lead to confounding factors due to 
off-policy data or
changed exploration behaviour, we utilise \acrfull{SL} here to provide initial 
evidence in a
more controlled setup.
We use the CIFAR-10 dataset for image classification \citep{krizhevsky2009learning} and artificially
inject non-stationarity. 

Our goal is to provide qualitative results on the impact of non-stationarity, not to obtain optimal
performance. 
We use a ResNet18 \citep{he2016deep} architecture, similar to those used
by \cite{espeholt2018impala} and \cite{cobbe2019leveraging}.
Parameters are updated using \gls{SGD} with momentum and, following standard practice in \gls{RL}, we use a constant learning rate and 
do not use batch normalisation.
Weight decay is used for regularisation. Hyper-parameters and more details can be found in \cref{sec:ap:sl}.

We train for a total of 2500 epochs. While the last 1500 epochs are trained on the 
full, unaltered dataset, we modify the training data in three different ways during the first 1000 epochs.
Test data is left unmodified throughout training.
While each modification is loosely motivated by the \gls{RL} setting, our goal is not to mimic it exactly
(which would be infeasible), nor to disentangle the contributions of different 
types of non-stationarity.
Instead, we aim to show that these effects reliably occur in the presence to 
various types of non-stationarity, and provide intuitions that can be brought 
into the RL setting in Section \ref{sec:iter}.

For the first modification, called \texttt{Dataset Size}, we initially train 
only on a small fraction of the
full dataset and gradually add more data points after 
each epoch, at a constant rate, until the full dataset is available after epoch 1000.
During the non-stationary phase, data points are reused multiple times to ensure the same 
number of network updates are made in each epoch.
For the modification \texttt{Wrong Labels} we replace all training labels by randomly drawn
incorrect ones. 
After each epoch, a constant number of these labels are replaced by their correct values.
Lastly, \texttt{Noisy Labels} is similar to \texttt{Wrong Labels}, but the incorrect labels are sampled uniformly at the 
start of each epoch. 
For both, all training labels are correct after epoch 1000.
While \texttt{Dataset Size} is inspired by the changing state distribution seen by an evolving
policy, \texttt{Wrong Labels} and \texttt{Noisy Labels} are motivated by the 
consistent bias or fluctuating errors a learned critic can introduce in the 
policy gradient estimate.

\begin{wrapfigure}{r}{0.67\textwidth}
   \centering
   \begin{subfigure}{0.49\columnwidth}
       \includegraphics[width=\linewidth]{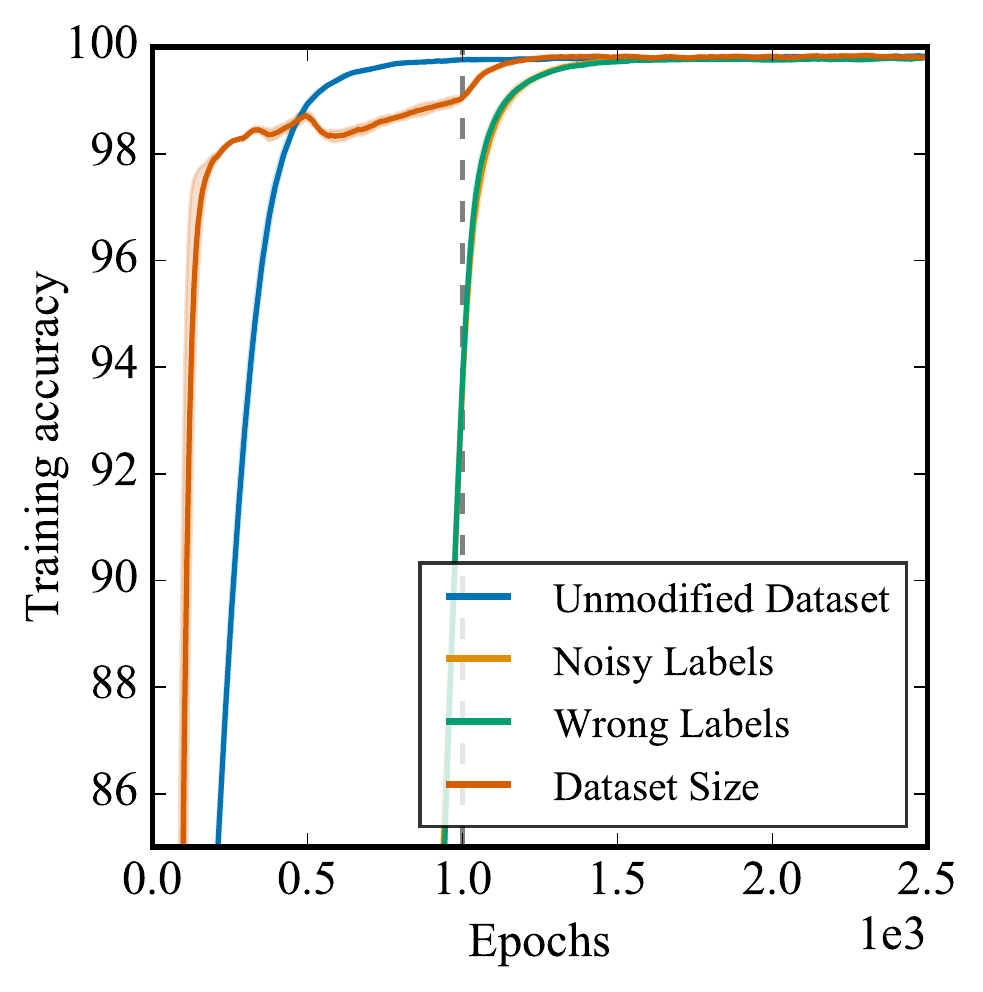}
   \end{subfigure}
   \begin{subfigure}{0.49\columnwidth}
       \includegraphics[width=\linewidth]{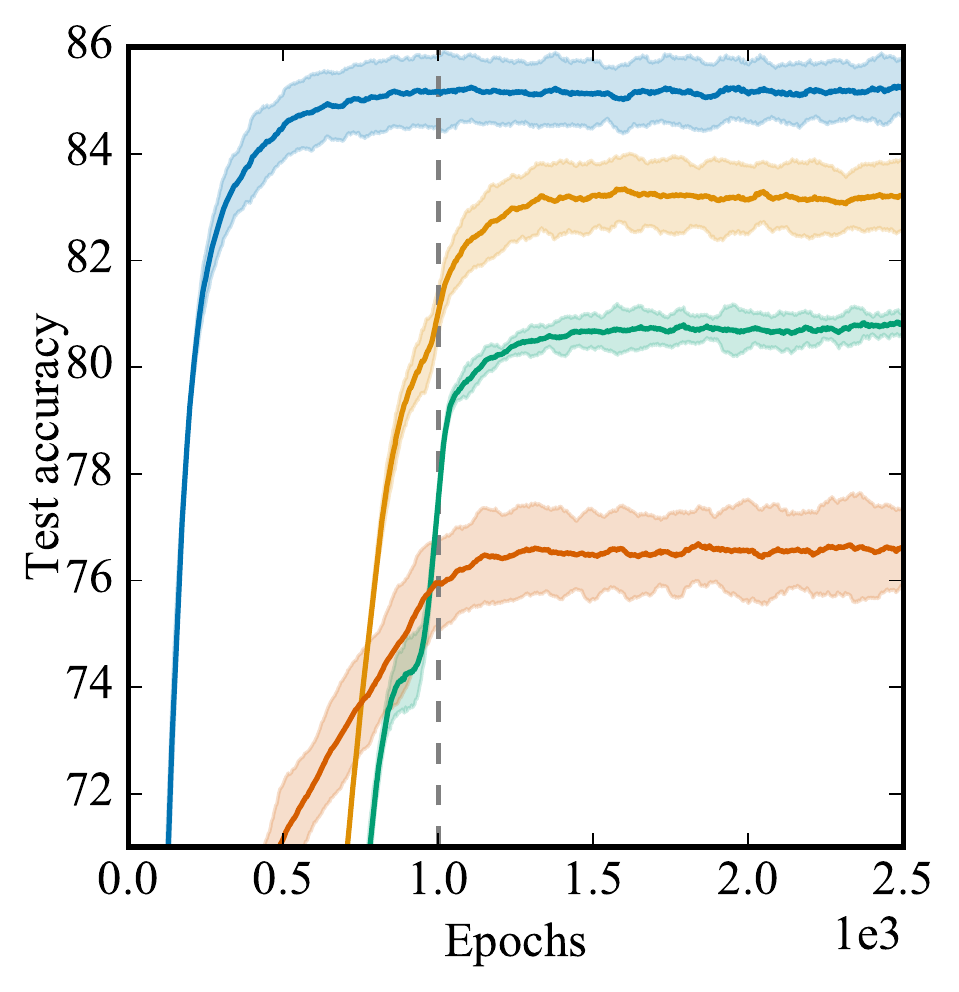}
   \end{subfigure}
   \caption{Accuracy on CIFAR-10 when the training data is
   non-stationary over the first 1000 epochs (dashed line). 
   The remaining epochs are trained on the full, unaltered training data. 
   Testing is performed on unaltered data throughout.
   While final training performance (left) is almost
   unaffected, test accuracy (right) is significantly reduced by initial, transient non-stationarity. }
   \label{fig:sl-nonstat}
\end{wrapfigure}

The results are shown in \cref{fig:sl-nonstat}. 
While the final training accuracy (left) is almost unaffected (see \cref{tab:sl} in the appendix for
exact results), all three non-stationary modifications significantly reduce the test accuracy (right).
The plateau in accuracy shows that this effect persists even after the models 
are further trained using the full dataset with correct labels:
non-stationarity early in training has a permanent effect on the learned 
representations and quality of generalisation.
These results indicate that the non-stationarity introduced by the gradual convergence of the policy
in \gls{RL} might similarly deteriorate the generalisation of the agent.
To overcome this, we propose \gls{ITER} in the next section.
The key insight enabling \gls{ITER} is that the observed negative effect is restricted to the test data,
whereas the predictions on the training data are unaffected and of high quality.

%% file: core/4_iter.tex
\section{ITER}
\vspace{-0.5em}
\label{sec:iter}
In \cref{sec:nonstat}, we have seen evidence that the
non-stationarity which is present in many deep \gls{RL} algorithms might lead to impaired
generalisation on held-out test environments.
To mitigate this and improve generalisation to previously unseen states, we propose \acrfull{ITER}:
instead of updating a single agent model throughout the entire training process,
\gls{ITER} learns a sequence of models, each of which is exposed to less non-stationarity during its
training.
As we will show in \cref{sec:experiments}, this improves generalisation.
\gls{ITER} can be applied on top of a wide range of base \gls{RL} training
methods.
For simplicity, we focus in the following exposition on actor-critic
methods and use \gls{PPO} \citep{schulman2017proximal} for the experimental evaluation.

The underlying insight behind \gls{ITER} is that at any point during \gls{RL} training
the latent representation of our current agent network might be significantly damaged by
non-stationarity, but its outputs on the training data are comparatively unaffected (see
\cref{fig:sl-nonstat}).
Consequently, \gls{ITER} aims to periodically replace the current agent network, the `teacher', by a `student'
network which was freshly initialised and trained to mimic the teacher on the current training data.
Because this re-learning and replacement step can be performed on stationary data,
it allows us to re-learn a policy that matches performance on the
training data but generalises better to novel test environments.

\gls{ITER} begins with an initial policy $\pi^{(0)}$ and value function
$V^{(0)}$ and then repeats the following steps, starting with iteration $k=0$.
\begin{packenum}
	\item Use the base \gls{RL} algorithm to train $\pi^{(k)}$ and $V^{(k)}$.
	\item Initialise new \emph{student} networks for $\pi^{(k+1)}$ and
	$V^{(k+1)}$.
	We refer to the current policy $\pi^{(k)}$ and value function
	$V^{\scriptscriptstyle (k)}$ as the \emph{teacher}.
	\item Distill the teacher into the student. This
	phase is discussed in more detail in \cref{sec:distillation}.
	\item The student replaces the teacher: $\pi^{(k)}$ and $V^{(k)}$ can be
	discarded.
	\item Increment $k$ and return to step 1. Repeat as
	many times as needed.
\end{packenum}

This results in alternating \gls{RL} training with distillation into a freshly
initialised student.
The \gls{RL} training phases continue to introduce
non-stationarity until the models converge, so we want to iterate the
process, repeating steps 1-4.
How often we do so depends on the environment and can be chosen as a hyper-parameter.
In practise we found the results to be quite robust to this choice and recommend, as general rule,
to iterate as often as possible within the limits outlined in \cref{sec:train_and_distill}.
There, we introduce a practical implementation of \gls{ITER} which re-uses data between steps 1 and
3 in order to not require additional samples from the environment.

\subsection{Distillation Loss}
\label{sec:distillation}

Our goal during the distillation phase (step 3) is to learn a new student policy
$\pi^{(k+1)}$ and value function $V^{(k+1)}$
that imitate the current teacher $(\pi^{(k)},
V^{(k)})$.
If the student and teacher share the same network architecture, the student
could of course imitate the teacher exactly by copying its parameters.
However, since the teacher was trained under non-stationarity, its
generalisation performance is likely degraded (see \cref{sec:nonstat}).
Consequently, we want to instead train a freshly initialised student network to
mimic the teacher's outputs for the available data, but learn a better
internal representation by training on a stationary data distribution collected by the
teacher $\pi^{(k)}$, i.e., $s,a\sim d_{\pi^{(k)}}(s) \pi^{(k)}(a|s)$.

The student, parameterised by $\theta_{k+1}$, is trained using a linear combination of four loss terms:
\begin{equation}
   \label{eq:loss}
   \mathcal{L}(\theta_{k+1}) = \alpha_\pi
   \mathcal{L}_\pi + \alpha_\text{V} \mathcal{L}_\text{V} +
   \mathcal{L}_\text{PG} +
   \lambda_\text{TD}\mathcal{L}_\text{TD}
\end{equation}
where $\lambda_{TD}$ is a fixed hyper-parameter and we linearly anneal
$\alpha_\pi$ and $\alpha_\text{V}$
from some fixed initial value to zero over the course of each distillation
phase.

$\mathcal{L}_\pi$ and $\mathcal{L}_V$ are supervised losses minimising the
disagreement between outputs of the student and the teacher:
\vspace{-3mm}
\begin{equation}
   \begin{split}
      \label{eq:supervised}
		\mathcal{L}_\pi(\theta_{k+1}) & =
		\mathbb{E}_{s\sim d_{\pi^{(k)}}}\Big[
			D_\text{KL}\big[\pi^{(k)}(\cdot|s) \,\big\|\,
			\pi^{(k+1)}(\cdot|s) \big] \Big] \,,
	\\
		\mathcal{L}_\text{V}(\theta_{k+1}) & =
			\mathbb{E}_{s\sim d_{\pi^{(k)}}}
			\Big[\Big(V^{(k)}(s) - V^{(k+1)}(s)\Big)^{\!2} \,\Big].
   \end{split}
\end{equation}
The additional terms $\mathcal{L}_\text{PG}$ and $\mathcal{L}_\text{TD}$ are
off-policy \gls{RL} objectives for updating the actor and critic:
\begin{equation}
   \label{eq:pg}
      \begin{split}
		   \mathcal{L}_\text{PG}(\theta_{k+1}) & = - \mathbb{E}_{s\sim d_{\pi^{(k)}}, a\sim \pi^{(k)}, s' \sim T(s,a)}
		   \Big[ \log \pi^{(k+1)}(a|s) \;
		   \bot \Big({\textstyle\frac{\pi^{(k+1)}(a|s)}{\pi^{(k)}(a|s)}} \,
			A^{(k+1)}(s,a,s') \Big) \Big] \,,
	   \\[0mm]
		   \mathcal{L}_\text{TD}(\theta_{k+1}) & =
		   \mathbb{E}_{s\sim d_{\pi^{(k)}}, a\sim \pi^{(k)}, s' \sim T(s,a)}
		   \Big[ \Big(A^{(k+1)}(s,a,s')\Big)^2 \,
		   	\bot{\textstyle\frac{\pi^{(k+1)}(a|s)}{\pi^{(k)}(a|s)}} \Big] \,,
      \end{split}
\end{equation}
where $A^{(k+1)}(s,a,s') = r(s, a) + \gamma	\bot V^{(k+1)}(s') - V^{(k+1)}(s)$
denotes the estimated advantage of choosing action $a$ and $\bot$ is a \texttt{stop-gradient}
operator, its operand is treated as a constant when taking derivatives of the objective.
In practice, the losses in \cref{eq:supervised}  remain nonzero during
distillation, potentially causing a drop in performance once the student
replaces the teacher.
The off-policy \gls{RL} losses in \cref{eq:pg} allow the student to already
take performance on the \gls{RL} task into account, reducing or eliminating
this performance drop.
We use \gls{PPO} losses to optimise \cref{eq:pg} in our experiments.

\subsection{Combining Training and Distillation}
\label{sec:train_and_distill}
To fully eliminate non-stationarity during the distillation step
we need to collect additional data from the environment using a fixed teacher policy.
However, this would slow down training by increasing the total number of samples required.
Here, to improve sample efficiency, we propose two practical implementations of \gls{ITER} which reuse
data between teacher and student:

	\emph{Sequential training:} Store the last $N$ transitions that
	were used to update the teacher in a dataset $\mathcal{D}$.
	During the distillation phase, draw batches from $\mathcal{D}$
	instead of collecting new data from the environment.
	While this does not introduce non-stationarity, it leads to evaluating
	the teacher on old, off-policy data, for which the quality of its outputs
	may be degraded.
	Furthermore, some of the data might be obsolete under the current state-distribution
	and we require additional memory to store $\mathcal{D}$.

	\emph{Parallel training:} Whenever the teacher is updated on a batch
	of data $\mathcal{B}$, also update the the student on the same batch.
	This approach introduces some non-stationarity as distillation is performed over multiple
	batches $\mathcal{B}$ while the teacher is changing.
	However, the teacher evolves much less over the course of the distillation
	phase than does a policy trained from scratch to achieve the same performance. In practise we
	found this to be a worthy trade-off.
	Advantages of this method are that no additional memory $\mathcal{D}$ is required, the
	teacher is only evaluated on data on which it is currently trained and updates to the student
	and teacher can be performed in parallel.

Both approaches perform similarly in our experimental validation.
We use parallel training for the main experiments due to the smaller
memory requirements, the ability to efficiently perform the
student distillation in parallel and because tuning the
hyper-parameter was significantly easier:
While tuning the size of $\mathcal{D}$ for sequential \gls{ITER} involves trading off overfitting
(for small $\mathcal{D}$) against off-policy data (for large $\mathcal{D}$), in parallel \gls{ITER}
the results were robust to the choice of how many batches $\mathcal{B}$ were used in the
distillation phase as long as some minimum number was surpassed.
Consequently, hyper-parameter tuning for parallel \gls{ITER} only involved increasing the length of
the distillation phase until no drop in student performance was observed when replacing the teacher.
We set $\alpha_\pi=1$ and $\alpha_V=0.5$ as initial values without further tuning as preliminary experiments showed no
impact within reasonable ranges.


%% file: core/5_experiments.tex
\section{Experiments}
\vspace{-0.5em}
\label{sec:experiments}

In the following, we evaluate \gls{ITER} on \emph{Multiroom} \citep{gym_minigrid}
and on several environments from the \emph{ProcGen} \citep{cobbe2019leveraging} benchmark which was
specifically designed to measure generalisation by introducing separate test- and training levels.
We also provide ablation studies showing that parallel and sequential training
perform comparably, and that the loss terms \cref{eq:pg} in \cref{eq:loss} are beneficial.
We find that \gls{ITER} improves generalisation, which also supports our hypothesis about the
negative impact of transient non-stationarity in \gls{RL}.
Lastly, we re-visit the \gls{SL} setting from
\cref{sec:nonstat} and perform additional experiments leading us to the formulation of the
\emph{legacy feature} hypothesis to explain the observed effects.

\subsection{Experimental Results on Multiroom}
\label{sec:mr}

\begin{figure}[t]
\vspace{-0.9em}
   \centering
   \begin{subfigure}{0.32\columnwidth}
       \centering
       \includegraphics[width=\linewidth]{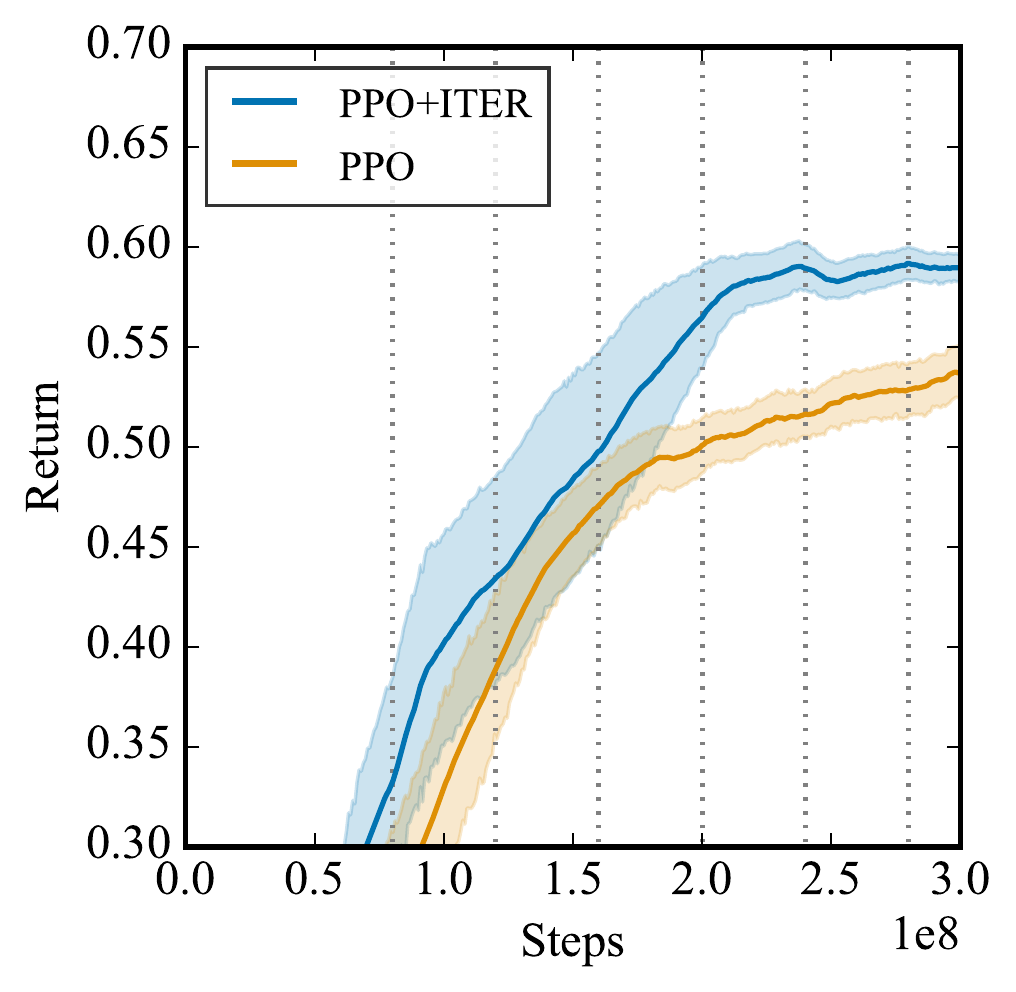}
       \caption{Multiroom}
   \end{subfigure}
   \begin{subfigure}{0.32\columnwidth}
       \includegraphics[width=\linewidth]{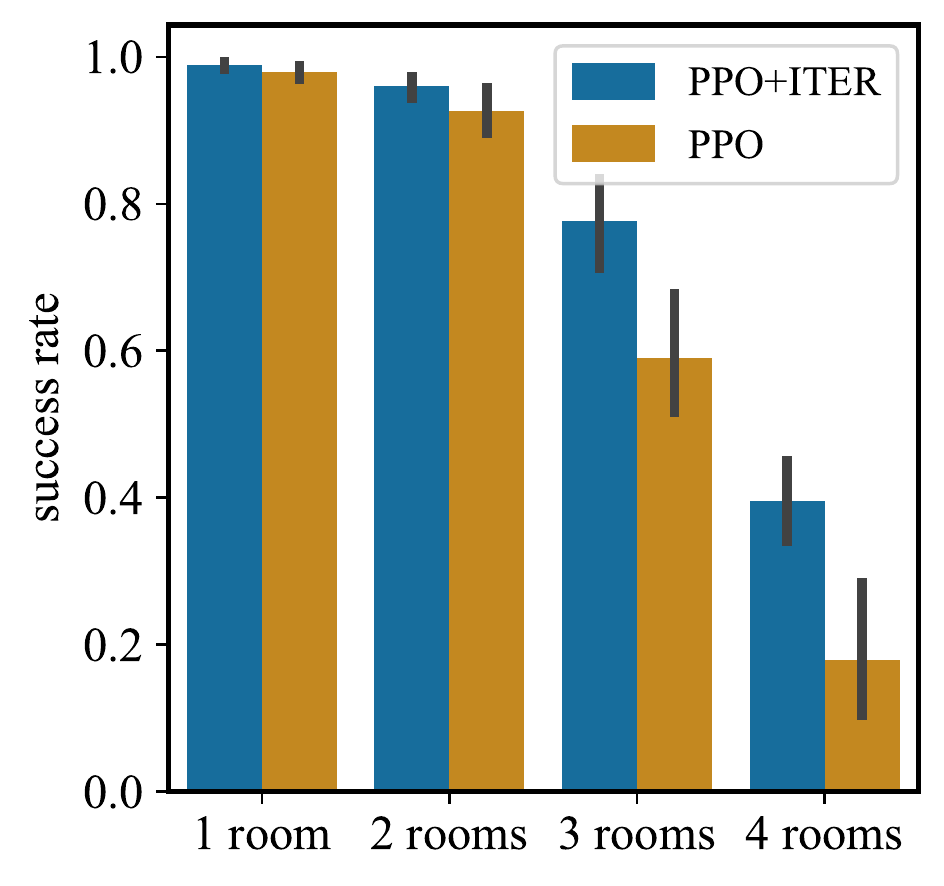}
       \caption{Multiroom, by layout}
   \end{subfigure}
   \begin{subfigure}{0.32\columnwidth}
       \includegraphics[width=\linewidth]{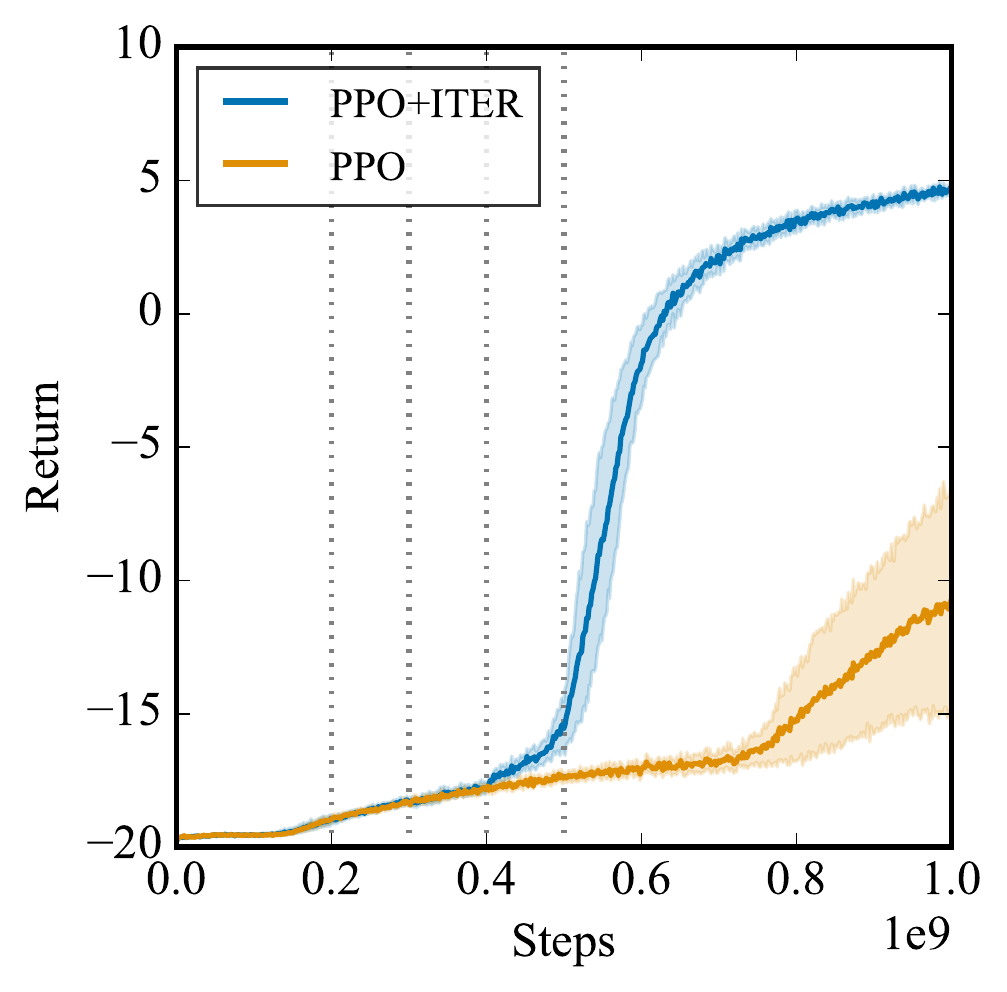}
       \caption{Boxoban}
   \end{subfigure}
   \caption{Evaluation on \emph{Multiroom} and \emph{Boxoban}.
   Shown are mean and standard error over twelve seeds.
   \emph{Left:} Return for \gls{PPO} with and without \gls{ITER} on \emph{Multiroom}.
   Dotted lines indicate when the network was replaced by a new student.
   \emph{Middle:} Evaluation on layouts with a fixed number of rooms; training is still
   with a random number of rooms. \gls{ITER}'s advantage is more pronounced for harder levels.
   \emph{Right:} Return on Boxoban.
   }
   \label{fig:results_multiroom}
\end{figure}

First, we evaluate \gls{ITER} on the \emph{Multiroom} environment.
The agent's task is to traverse a sequence of rooms to reach the goal (see
\cref{fig:sl_f} for example layout) as quickly as possible.
It can take discrete actions to rotate 90° in either direction, move forward, and open or close the doors
between rooms.
The observation contains the full grid, one pixel per square.
Object type, including empty space and walls, as well as any object status, like direction, are encoded
in the three `colour' channels.
For each episode, a new layout is generated by randomly placing between one and
four connected rooms on the grid.
The agent is always initialised in the room furthest from the goal.
This randomness favours agents that are better at generalising between layouts as memorisation is
impossible due to the high number of feasible layouts.
The results are shown in \cref{fig:results_multiroom}: Using \gls{ITER} on top of \gls{PPO}
increases performance.
The performance difference is more pronounced for layouts with more rooms,
possibly because such layouts are harder and likely only solved later in training, at which
point negative effects due to prior non-stationarity in training are more pronounced.

\subsection{Experimental Results on Boxoban}

We also evaluate \gls{ITER} on \emph{Boxoban} \citep{boxobanlevels,SchraderSokoban2018}\footnote{
Our simplified \texttt{Boxoban-Train-v0} also allows pulling boxes to reduce computational costs}.
See \cref{fig:ap_boxoban} for an example.
Similarly to \emph{Multiroom} the observation contains the full grid, one pixel per square. Object
types are encoded by colour.
Again, a new layout is generated at the beginning of each new episode, favouring agents that can
generalise well between states.
Actions allow to push, pull or move in all four cardinal directions, or do nothing.
The goal of the agent is to position the four boxes, which can be pushed or pulled, on the four available
targets. Walls prevent movement for both the agent and the boxes.
Positive rewards are provided for positioning a box on a target ($r_b=1$) and successfully solving each level
($r_l=10$). A small negative reward per time-step ($r_s=-0.1$) encourages fast solutions.
As shown in \cref{fig:results_multiroom}, \gls{ITER} learns much faster.
We provide additional results and examples for wrongly chosen distillation lengths in
\cref{fig:ap_boxoban}.
Note that both for \emph{Multiroom} and \emph{Boxoban} we train and test on the same (very large)
set of possible layout configurations, so the main expected advantage of \gls{ITER} is a more
sample efficient training due to better generalisation.
In the next section, we will evaluate the agent on previously unseen environments, directly
measuring its generalisation performance.

\subsection{Experimental Results on ProcGen}
\label{sec:procgen}

\begin{figure}[t]
\vspace{-0.9em}
   \centering
   \begin{subfigure}{0.32\columnwidth}
       \includegraphics[width=\linewidth]{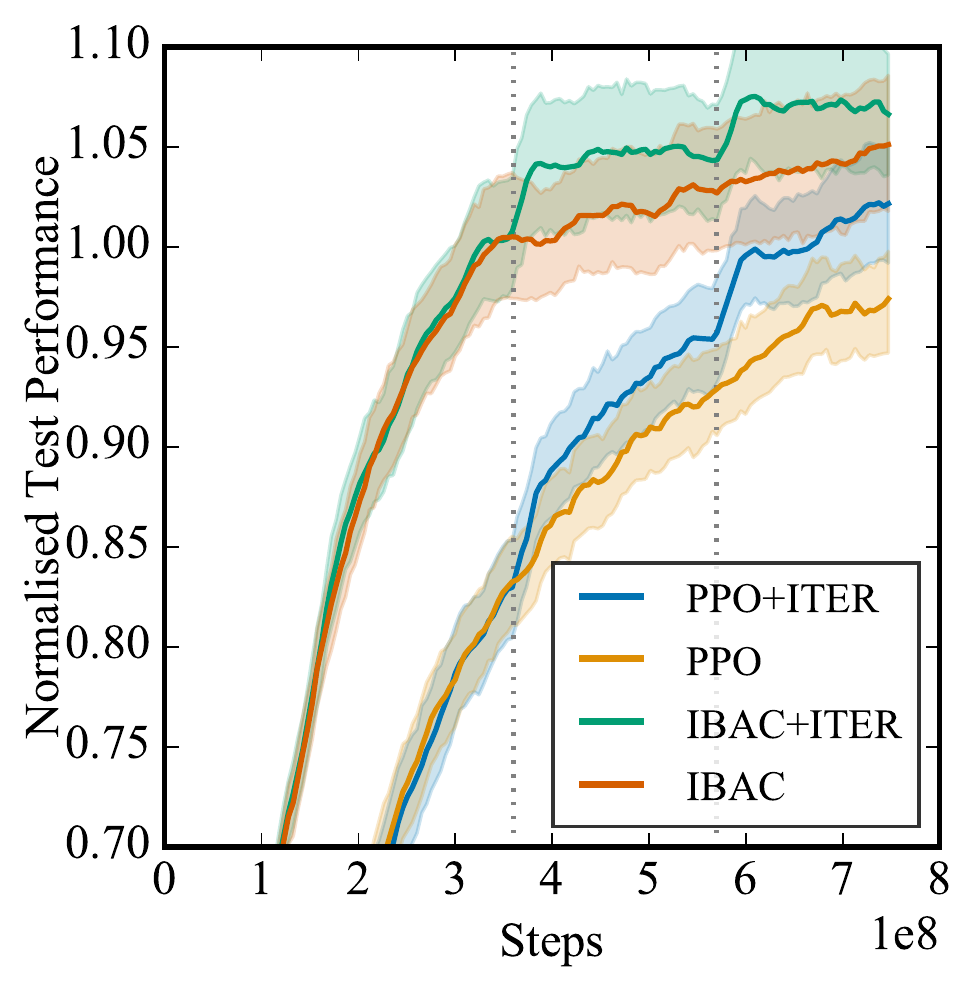}
   \end{subfigure}
   \hfill
   \begin{subfigure}{0.32\columnwidth}
       \centering
       \includegraphics[width=\linewidth]{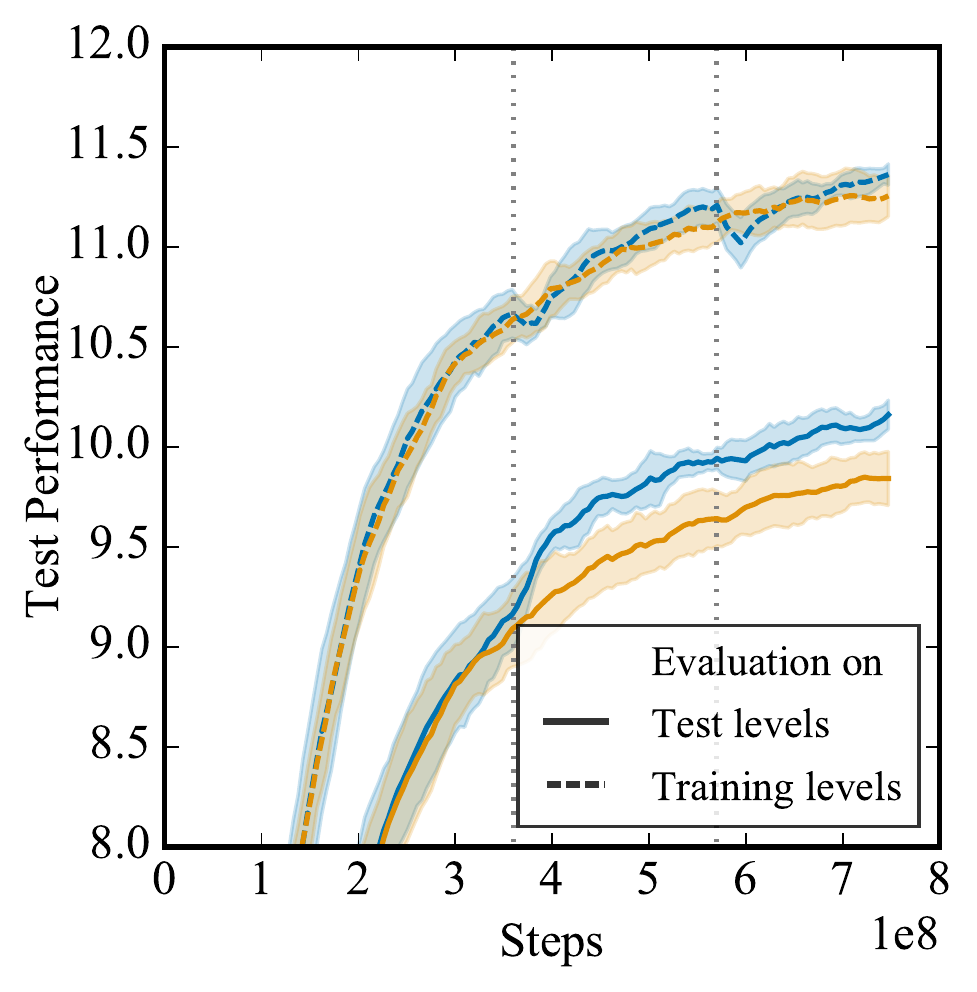}
   \end{subfigure}
   \hfill
   \begin{subfigure}{0.32\columnwidth}
       \centering
       \includegraphics[width=\linewidth]{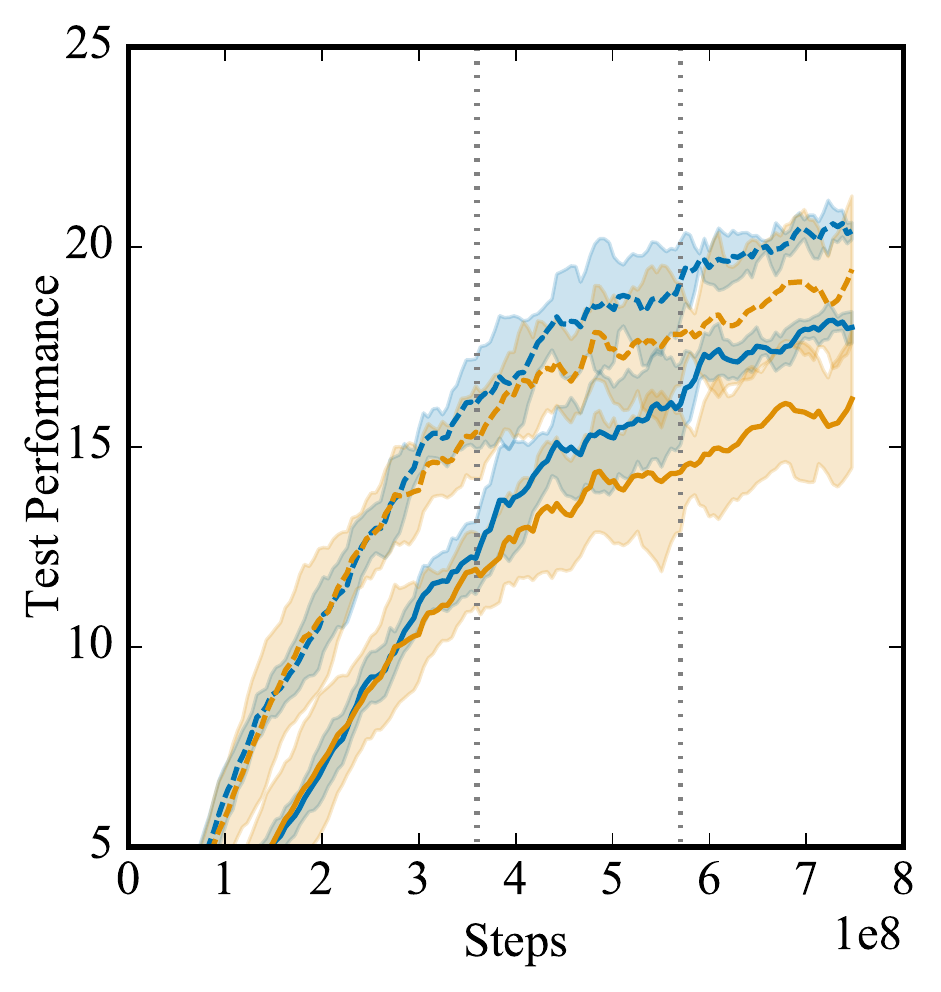}
   \end{subfigure}
   \caption{Evaluation on \emph{ProcGen}. Dashed lines indicate replacing the teacher.
   \emph{Left:} Test performance averaged over six environments (\emph{StarPilot},
   \emph{Dodgeball}, \emph{Climber}, \emph{Ninja}, \emph{Fruitbot} and \emph{BigFish}).
   Shown are mean and standard error over all 30 runs (five per environment).
   Results are normalised by the final test-performance of the \gls{PPO} baseline on each respective
   environment to make them comparable.
   We also compare against the previous state of the art method \textsc{IBAC-SNI} \citep{igl2019generalization}.
   \emph{Middle:} Evaluation on \emph{Climber}. \gls{ITER} improves test performance
   without improving training, supporting our claim that \gls{ITER} improves the
   latent representation of the agent.
   \emph{Right:} Evaluation on \emph{BigFish}. On some environments, \gls{ITER} improves both train- and test-
   performance.
    \vspace{-1em}
   }
   \label{fig:results_procgen}
\end{figure}
%

\begin{figure}[t]
	\centering
	\begin{subfigure}{0.33\columnwidth}
		\includegraphics[width=\linewidth]{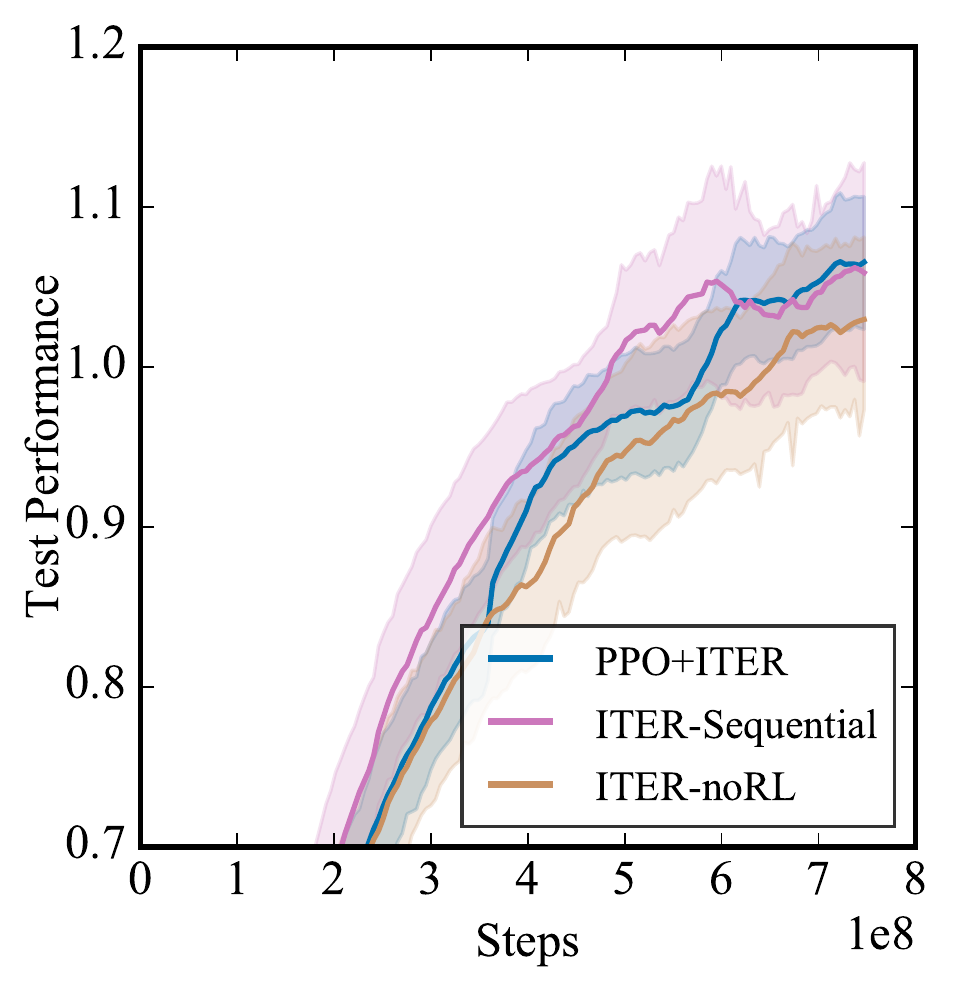}
	\end{subfigure}
    \begin{subfigure}{0.64\columnwidth}
        \centering
        \includegraphics[width=0.8\linewidth]{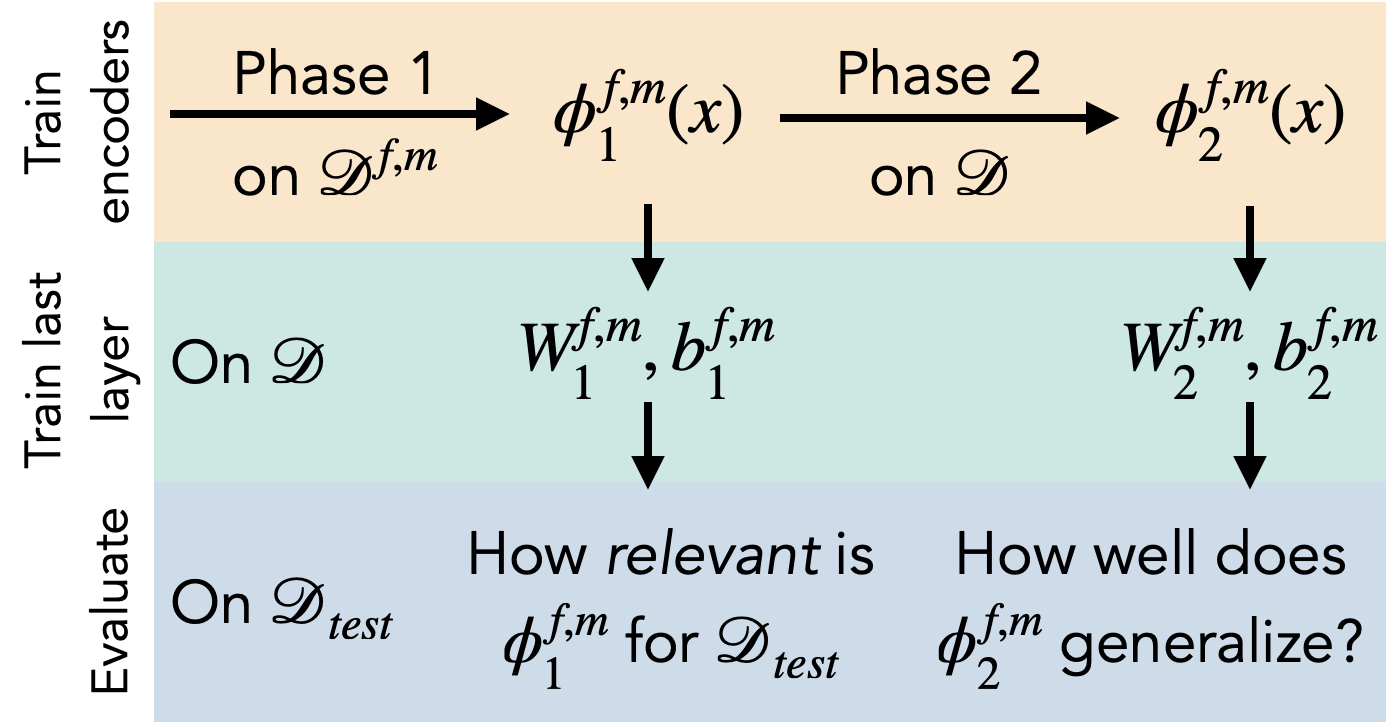}
	\end{subfigure}
	\caption{\emph{Left:} Ablation studies with sequential \gls{ITER} and \gls{ITER} without
        terms $\mathcal{L}_{\text{PG}}$ and $\mathcal{L}_{\text{TD}}$ (\cref{eq:pg}).
        \emph{Right:} Schematic depiction of training setup for \cref{fig:sl} (middle and right). More details are
        given in \cref{sec:exp:sl}.
        $\mathcal{D}$ is the unmodified CIFAR-10 training data while for $\mathcal{D}^{f,m}$
        modification $m\in\{\texttt{Noisy Labels},\texttt{Wrong Labels}, \texttt{Dataset Size}\}$ is
        applied to the fraction $(1-f)$ of all data-points. In this two phase training setup, we
        first train on $\mathcal{D}^{f,m}$ during phase 1 and continue on $\mathcal{D}$ during phase 2. A linear predictor
        parameterised by $(W^{f,m}_i,b^{f,m}_i)$ is trained on $\mathcal{D}$ after each phase $i$,
        while holding the encoder $\phi_i^{f,m}(x)$ fixed.
        Evaluation of the resulting classifiers is performed on the original test data.
        Classifier $i=1$ measures the relevance of the legacy features while classifier $i=2$
        measure the final generalisation performance.
        }
	\label{fig:ablation}
\end{figure}
Next, we evaluate \gls{ITER} on several environments from the \emph{ProcGen} benchmark.
We follow the evaluation protocol proposed in \citep{cobbe2019leveraging}: for each environment, we train on 500 randomly
generated level layouts and test on additional, previously unseen levels.
Due to computational constraints, we consider a subset of five environments.
We chose \emph{StarPilot}, \emph{Dodgeball}, \emph{Climber}, \emph{Ninja}, \emph{Fruitbot} and \emph{BigFish} based
on the results presented in \citep{cobbe2019leveraging} as they showed baseline generalisation
performance better than a random policy, but with a large enough generalisation gap.
\gls{ITER} improves performance for both \gls{PPO} and \gls{IBAC} with selective noise injection
\citep{igl2019generalization}.
Results are presented in \cref{fig:results_procgen} and more individual plots, including performance
on training levels, can be found in the appendix.
In \cref{fig:ablation} we show in ablations that both the parallel and sequential implementations of \gls{ITER}
perform comparably, while not using the off-policy RL terms $\mathcal{L}_{\text{PG}}$ and
$\mathcal{L}_{\text{TD}}$  in \cref{eq:loss} decreases performance.

Similarly to previous literature \citep{cobbe2018quantifying,igl2019generalization}, we found that weight decay improves performance
and apply it to all algorithms evaluated on \emph{ProcGen}.
Our results show that the negative effects of non-stationarity cannot easily be
avoided through standard network regularisation: we can improve test returns through \gls{ITER}
despite regularisation with weight decay and \gls{IBAC}, both shown to be among the most effective
regularisation methods on this benchmark \citep{igl2019generalization}.

In the previous two sections we have shown the effectiveness of \gls{ITER} in improving generalisation of \gls{RL}
agents. Because the main mechanism of \gls{ITER} is in reducing the non-stationarity in the training
data which is used to train the agent, this result further supports that such transient
non-stationarity is detrimental to generalisation in \gls{RL}.

\subsection{Supervised Learning on CIFAR-10}
\label{sec:exp:sl}
\begin{figure}[ht]
   \centering
   \begin{subfigure}{0.28\columnwidth}
       \includegraphics[width=\linewidth]{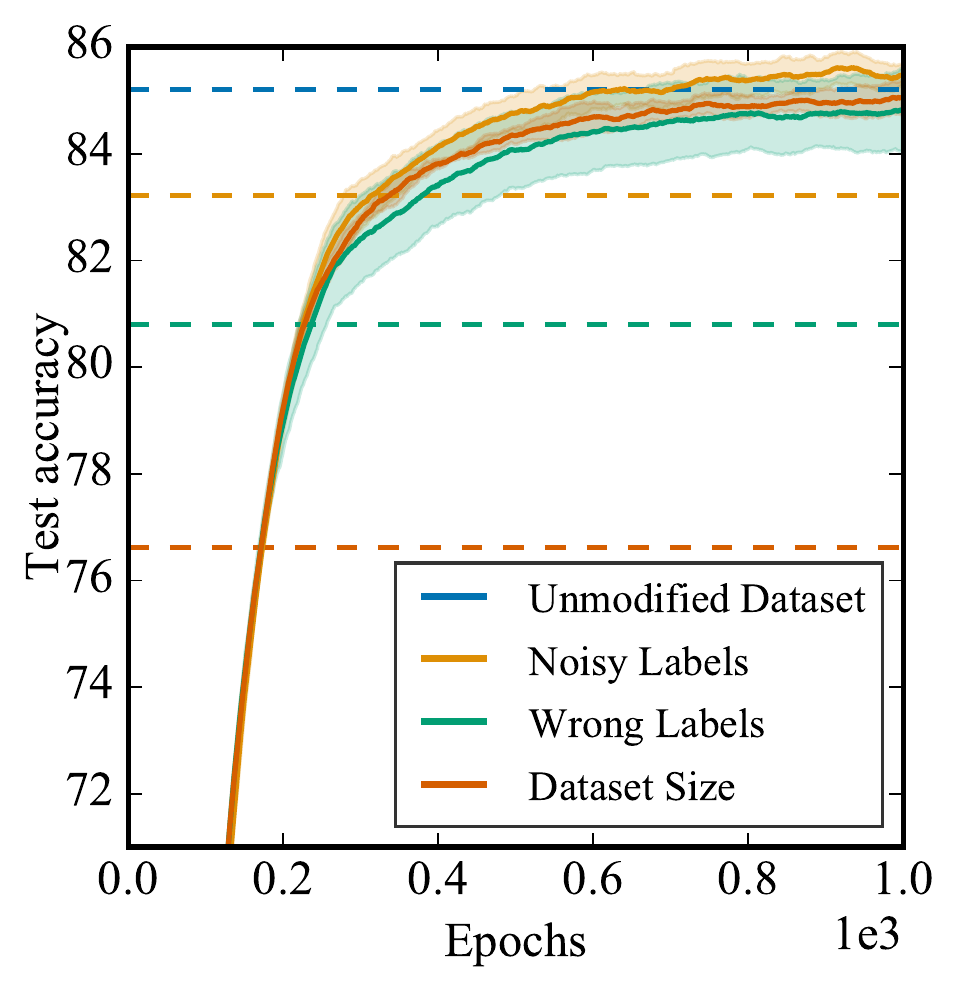}
   \end{subfigure}
   \hfill
   \begin{subfigure}{0.40\columnwidth}
       \includegraphics[width=\linewidth]{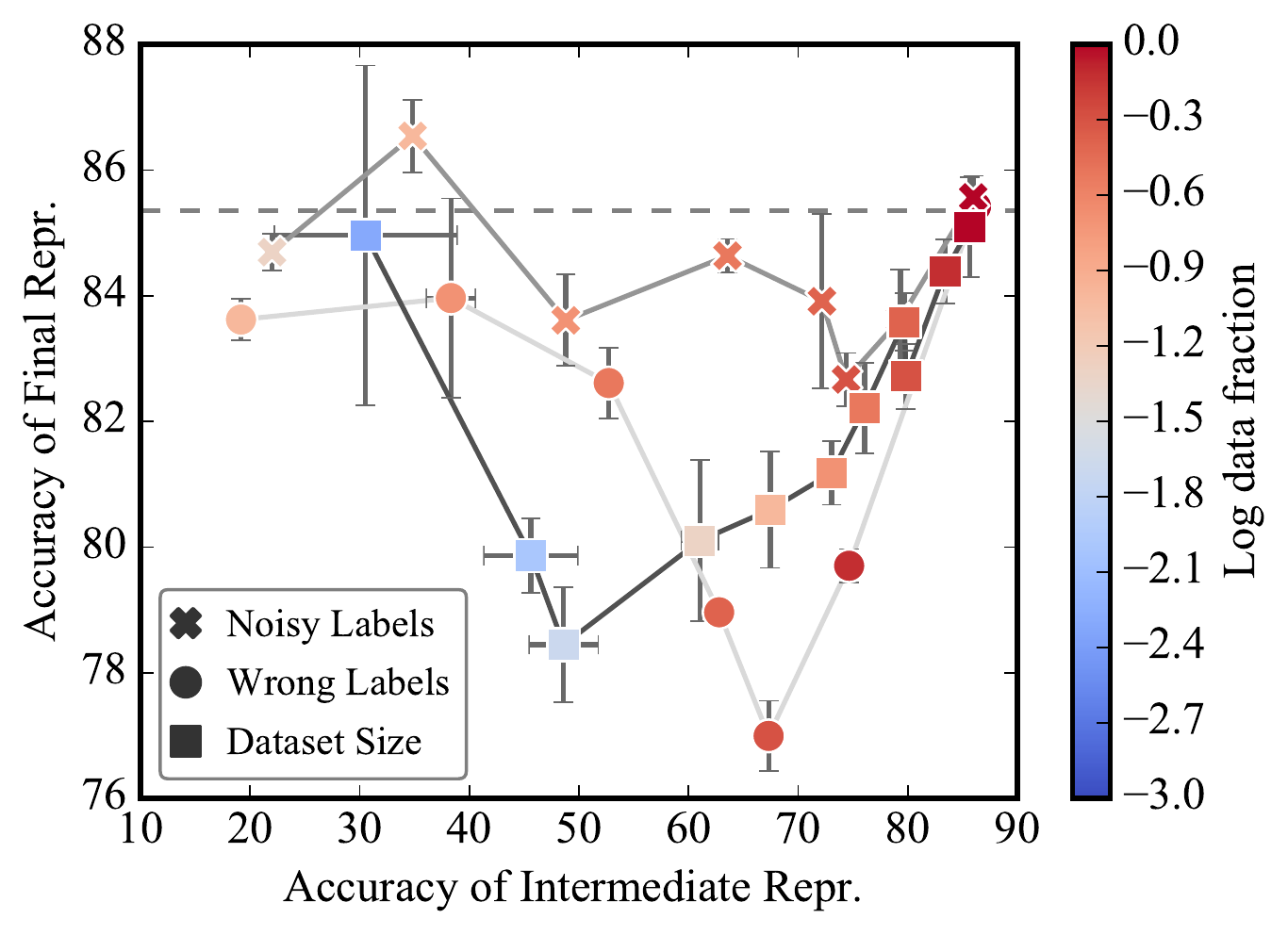}
   \end{subfigure}
   \hfill
   \begin{subfigure}{0.28\columnwidth}
       \includegraphics[width=\linewidth]{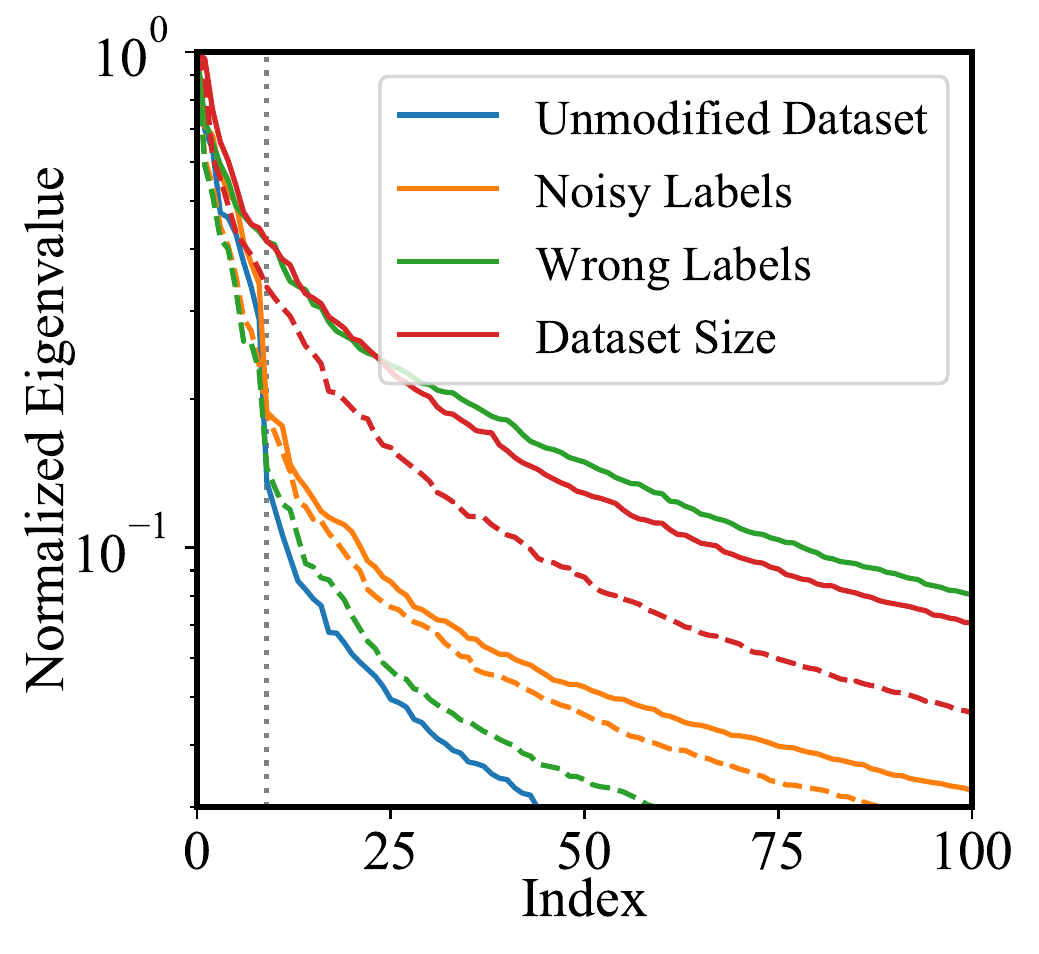}
   \end{subfigure}
   \caption{\emph{Left:} Test accuracy of students (solid lines) that only learn to mimic the
   behaviour of poorly generalising teachers in \cref{fig:sl-nonstat} (dashed lines).
   \emph{Middle:} Final test accuracy of networks trained consecutively on two different
   datasets.
   The $x$-axis shows the accuracy of using encoders trained on the first
   dataset, retraining only the last layer on the second:
   nearly useless earlier representations impact future learning
   much less than slightly sub-optimal ones.
   Markers indicate modifications to first dataset; colours indicate the fraction of unmodified
   data points $f$.
   Dashed line shows accuracy for $f=1$.
   \emph{Right:} Singular values of feature matrix $\Phi$, normalised by the largest SV. Solid lines show
   intermediate values of $f$ with low test accuracy, dashed lines small values of $f$ with
   higher accuracy. More plots can be found in the appendix.
   \vspace{-1em}
   }
   \label{fig:sl}
\end{figure}


In this section, we aim to further understand the mechanism by which
non-stationarity impacts generalisation by revisiting the easily controlled
\gls{SL} setting presented in \cref{sec:nonstat}.

First, we confirm that, like with ITER for \gls{RL}, we can improve
generalisation while only learning to mimic the outputs of a poorly generalising
teacher for the training data.
We train the teacher using the same setup as in \cref{sec:nonstat}.
In a second step, we train a freshly initialised student for 1000 epochs to fit the
argmax predictions of this teacher on the training data, i.e., the
true labels in the training data are unused.
Test accuracy is still measured using the true labels.
The results of this distillation phase are shown in \cref{fig:sl} (left).
The student (solid lines) recovers the test accuracy
achieved by stationary training, compared to the poor asymptotic
teacher performance (dashed lines) from \cref{fig:sl-nonstat}.
This confirms that the teacher's outputs on the training data are
suitable targets for distillation.

Second, we aim to better understand why non-stationarity affects the generalisation
performance.
To do so, we investigate the latent representation of the network.
We view all but the last layer as the encoder, producing the latent \emph{representation}
$\phi(x)\in\mathbb{R}^p$ for input $x$, on which
the classification is linear: $y=\mathrm{softmax}(W\phi(x) + b)$ with $W\in\mathbb{R}^{|C|\times p}$ denoting a weight matrix
and $b\in\mathbb{R}^{|C|}$ denoting a bias vector.
By \emph{features} we refer not to the representation $\phi(x)$ as a whole, but to aspects
of the input to which the encoder learned to pay attention and which therefore can
impact the latent representation \citep{kriegeskorte2013representational}.
More quantitatively, we can define the representation matrix $\Phi\in\mathbb{R}^{N\times p}$, consisting
of the latent representations of all $N=10000$ test data points.
Performing \gls{SVD} on $\Phi$ yields mutually orthogonal directions (the
right-singular vectors) in which the latent representations of the various inputs are
different from one another.
We can see each such direction as corresponding to one feature, with the corresponding singular value expressing
its strength, i.e., how strongly it impacts the latent representation.

Our hypothesis is that under a non-stationary data distribution,
the encoder is more likely to reuse previously learned features (as these are already available)
instead of learning new features from scratch.
If these old (or `legacy') features generalise worse on the new data, for example because they are
overfit to a smaller or less diverse dataset, this in turn deteriorates generalisation permanently if they are not replaced.
This leads to two predictions: First, the observed drop in generalisation should only occur if the
previously learned features are relevant for the new task, but suboptimal.
If they are irrelevant, they will not be reused.
If they are optimal, they do not negatively impact generalisation.
Second, we expect the final network to rely on more features in its latent representation if it is reusing suboptimal
features: because these are not as general, more features are required to discriminate
between all inputs.

To experimentally evaluate both predictions, we simplify the experimental setting to two phases (see
\cref{fig:ablation} for a schematic depiction of the training setup).
The first training phase uses a stationary, but modified, dataset
$\Dfm$, and the second phase uses the full, unmodified, training dataset $\mathcal D$.
To generate $\Dfm$, we use the same modifications as before, $m
\in \{\texttt{Noisy Labels},\texttt{Wrong Labels}, \texttt{Dataset Size}\}$,
but instead of annealing the fraction of correct data points $f$ from 0 to 1 as in
\cref{sec:nonstat}, it is fixed at a certain value.
Changing this value $f$ allows us to tune the relevance of the features learned on
$\Dfm$ (also see \cref{fig:sl_f}).
In this setup the only non-stationarity is the change in data from phase 1 to phase 2.
We first train the network for 700 epochs on $\Dfm$,
which yields an intermediate representation $\phi^{f,m}_{inter}(x)$, followed by another 800 epochs
on $\mathcal{D}$ yielding the final representation $\phi^{f,m}_{final}(x)$ (see
\cref{fig:sl_detailed_training} for training curves).

To test our first hypothesis, we want to measure how relevant the representation $\phii$ is for the
final data distribution and how well $\phif$ generalises.
We train a linear predictor for each \emph{fixed} representation on the \emph{full} dataset $\mathcal{D}$.\footnote{
	The linear predictor is $y=\sigma(W\phi_{f,m}(x)+b)$,
	where $\sigma$ is the softmax function and $x$ the input image.
}
The test accuracy of the classifier based on $\phii$ measures how well we can perform on
$\mathcal{D}$ with features learned on $\Dfm$, i.e., their \emph{relevance}.
The test accuracy of the classifier based on $\phi_{final}^{f,m}$ measures how well the final
 network was able to recover
from the initial bias and learn to generalise well despite non-stationarity.

In \cref{fig:sl} (middle), we plot the accuracy of the intermediate predictor (i.e., the relevance) on the
$x$-axis and of the final generalisation performance on the $y$-axis.
Each point corresponds to one value of $f\in(0, 1]$, shown as colour, and one modification type $m$ as indicated by the
marker shape.
By changing $f$ from 0 to 1 (i.e., from blue to dark red) we can increase the relevance of the
intermediate features from nearly irrelevant to optimal on $\mathcal{D}$.
Interestingly, an almost useless intermediate representation (30\% on the $x$-axis) does not impede
the final performance much, while relevant but suboptimal intermediate features (around 60\% on the
$x$-axis) lead to a marked drop in performance.
This supports our first hypothesis.
The strong final performance for $f \to 0$ (i.e., for low relevance) also rules out decreased
network flexibility, for example due to dead neurons (for ReLUs) or
saturation (for tanh), as the main driver of reduced generalisation.

Our second prediction is that relevant, but suboptimal features in $\phi_{inter}^{f,m}$ should lead
to the usage of more features in $\phi_{final}^{f,m}$ compared to irrelevant or optimal
features.
To test this prediction, we plot in \Cref{fig:sl} (right) the singular values (SVs) of the representation matrix
$\Phi\in\mathbb{R}^{N\times p}$ as defined above.
We plot the values for the smallest values of $f$ (dashed lines, ``irrelevant features'') and for
intermediate values of $f$ (solid lines, `relevant but sub-optimal features').
The blue line corresponds to optimal features.
The tails of the singular values are heavier for intermediate values of $f$,
indicating that the network is relying on more features in those cases,
supporting our second prediction.

%% file: core/6_relatedWork.tex
\section{Related Work}
\vspace{-0.5em}

Knowledge distillation \citep{hinton2015distilling} with identical teacher and student architectures
has been shown to improve test accuracy \citep{furlanello2018born}, even in the absence of non-stationarities in the data.
This improvement has been attributed to the ease of predicting the output distribution of
the teacher compared to the original `hard' labels \citep{mobahi2020self,gotmare2018a}.
While we apply such `soft' distillation for \gls{ITER} on \gls{RL}, we use `hard' labels in our \gls{SL}
experiments.

Policy distillation has been applied to \gls{RL} \citep{czarnecki2019distilling}, for example for
multi-task learning and compression \citep{teh2017distral,rusu2015policy, parisotto2016actor},
imitation learning \citep{ross2011reduction}, or faster training \citep{schmitt2018kickstarting,ghosh2018divideandconquer}.
Closer to \gls{ITER}, \citet{czarnecki2018mix} use policy distillation to learn
a sequence of agents.
However, their Mix \& Match algorithm solves tasks of growing complexity,
for example, to grow the action space of the agent,
not to tackle generalisation or non-stationarity.

While the topic of non-stationarity is central to the area of continual learning (see
\citep{parisi2019continual} for a recent review), the field is primarily concerned with preventing
catastrophic forgetting \citep{french1999catastrophic} when the
environment or task changes during training \citep{li2017learning,schwarz2018progress}.
For non-stationary environments during agent deployment, the approach is
typically to detect such changes and respond accordingly \citep{choi2000environment,da2006dealing,doya2002multiple}.
By contrast, we assume a stationary environment and investigate the impact of transient non-stationarity,
for example induced by an improving policy.
We also show that intentionally forgetting the representation, but not the
learned outputs, can improve generalisation in this case.

Neural networks are used in deep \gls{RL} to allow generalisation across similar states
\citep{sutton2011reinforcement}.
Several possibilities have been proposed to further improve generalisation, including
to provide more diverse training environments \citep{tobin2017domain}, inject
noise into the environment \citep{stulp2011learning,Lee2020Network}, incorporate inductive biases in the
architecture \citep{kansky2017schema}, or regularise the network
\citep{cobbe2018quantifying,igl2019generalization,liu2019regularization}.
While regularisation reduces overfitting, we show in our experiments
that this is insufficient to counter the negative effects of
non-stationarity, and that \gls{ITER} can be complementary to other types of
regularisation.

%% file: core/7_conclusions.tex
\section{Conclusion}
\vspace{-0.5em}

In this work, we investigate the impact of non-stationarity on the generalisation performance of
trained \gls{RL} agents.
First, in several \gls{SL} experiments on the CIFAR-10 dataset, we confirm that non-stationarity can
considerably degrade test performance while leaving training performance nearly unchanged.
To explain this effect, we propose and experimentally support the \emph{legacy feature} hypothesis
that networks exhibit a tendency to adopt, rather then forget, features learned earlier during
training if they are sufficiently relevant, though not necessarily optimal, for the new data.
We also show that self-distillation, even without using the true training labels, improves performance on the test-data.

Many deep \gls{RL} algorithms induce similar transient non-stationarity, for example due to a
gradually converging policy.
Consequently, to improve generalisation in deep \gls{RL}, we propose \gls{ITER} which reduces the
non-stationarity the agent networks experience during training.
Our experimental results on the \emph{Multiroom} and \emph{ProcGen} benchmarks
empirically support the benefits of \gls{ITER}, indicating that transient non-stationarity has a
negative impact in deep \gls{RL}.

%% file: core/9_appendix.tex
\section{Pseudo code}
\begin{algorithm}[h]
	\caption{Pseudo-Code for parallel ITER} \label{alg:pseudo_code}
	\textbf{Input} Length of initial RL training phase $t_{init}$, length of distillation phase $t_{distill}$ \\
    \textbf{Initialise} $k\leftarrow 0$, policy $\pi^{(k)}$, value function $V^{(k)}$ \\
    \BlankLine
    \emph{// Normal RL training at the beginning} \\
    \For{\upshape$t_{init} \text{ steps}$}{
        $\mathcal{B} \leftarrow \text{ collect trajectory data using } \pi^{(0)}$ \\
        Update $\pi^{(0)}$ and $V^{(0)}$ using standard RL method using $\mathcal{B}$ \\
    }
    \BlankLine
    \emph{// Combine further RL training of $\pi^{(k)},V^{(k)}$ with distillation of
        $\pi^{(k+1)},V^{(k+1)}$} \\
	\While{\upshape not converged}
	{
        \textbf{Initialise} student policy $\pi^{(k+1)}$ and value function $V^{(k+1)}$ \\
        \For{\upshape$t_{distill} \text{ steps}$}{
            $\alpha_V, \alpha_\pi \leftarrow \text{ linear annealing to 0 over } t_{distill} \text{ steps}$ \\
            $\mathcal{B} \leftarrow \text{ collect trajectory data using } \pi^{(k)}$ \\
            Update $\pi^{(k)}$ and $V^{(k)}$ with standard RL method using $\mathcal{B}$ \\
            Update $\pi^{(k+1)}$ and $V^{(k+1)}$ with \cref{eq:loss} using $\mathcal{B}$,
            $\alpha_V$,  $\alpha_\pi$, $\pi^{(k)}$ and $V^{(k)}$\\
        }
        \BlankLine
        \emph{// Housekeeping} \\
        Discard $\pi^{(k)}$ and $V^{(k)}$ \\
        Set $k \leftarrow k+1$ \\
    }
\end{algorithm}
\section{Supervised Learning}
\label{sec:ap:sl}

\begin{table}[h]
\begin{floatrow}\BottomFloatBoxes
    \ttabbox
    {\caption{Numerical values of results presented in \cref{fig:sl-nonstat}. The `Rel' column
    shows the error normalised by the error of the unmodified dataset.
    The error on the test-data deteriorates worse than on the training data, not only in absolute,
    but also relativ terms.}
    \label{tab:sl}}
    {
    \begin{tabular}{lllll}
            \toprule
            & \multicolumn{2}{l}{Training} & \multicolumn{2}{l}{Testing} \\
            & Error in \%         & Rel.   & Error in \%        & Rel.   \\
            \midrule
            Unmodified   & $0.17 \pm 0.09$     & 1.0    & $14.8 \pm 0.70$    & 1.0    \\
            Noisy Labels & $0.19 \pm 0.09$     & 1.13   & $16.8 \pm 0.70$    & 1.14   \\
            Wrong Labels & $ 0.20 \pm 0.08$    & 1.22   & $19.2 \pm 0.43$    & 1.30   \\
            Dataset Size & $ 0.18 \pm 0.08$    & 1.05   & $23.4 \pm 0.83$    & 1.58   \\
            \bottomrule
        \end{tabular}
    }
    \ttabbox
    {\caption{Hyper-parameters used in the supervised learning experiment on CIFAR-10}\label{tab:sl_params}}
    {
        \begin{tabular}{ll}
            \toprule
            Hyper-parameter     & Value  \\
            \midrule
            SGD: Learning rate & $\num{3e-4}$ \\
            SGD: Momentum      & $\num{0.9}$  \\
            SGD: Weight decay  & $\num{5e-4}$ \\
            \bottomrule
        \end{tabular}
    }
\end{floatrow}
\end{table}


\begin{figure}[ht]
   \centering
   \begin{subfigure}{0.45\columnwidth}
       \includegraphics[width=\linewidth]{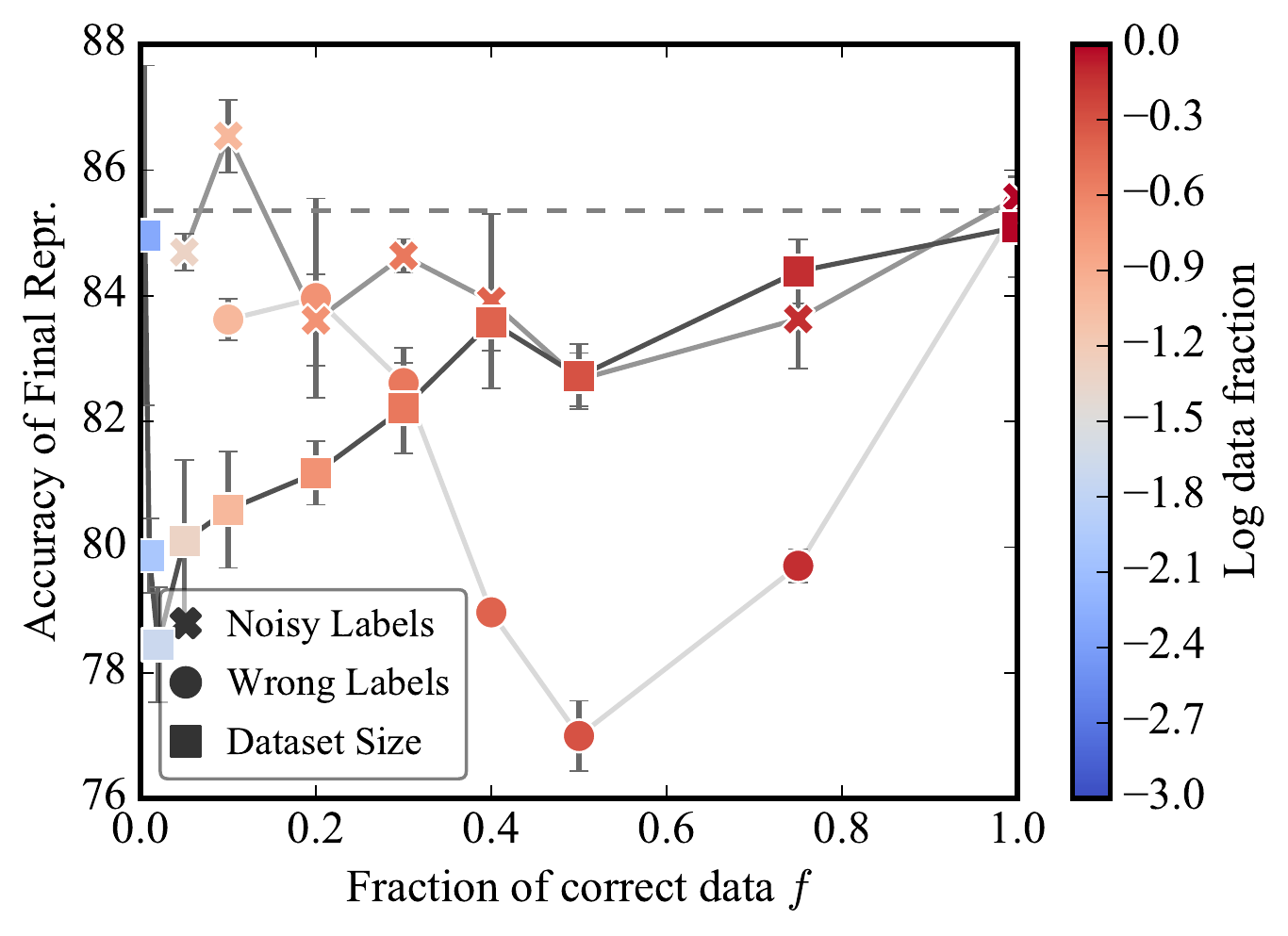}
   \end{subfigure}
   \begin{subfigure}{0.30\columnwidth}
       \includegraphics[width=\linewidth]{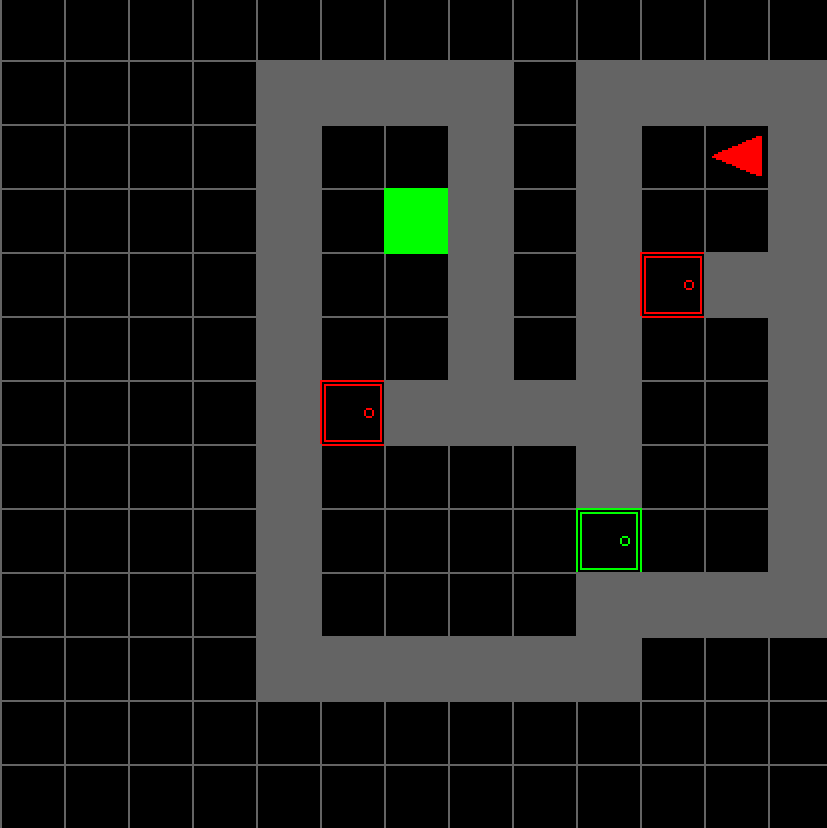}
   \end{subfigure}
   \caption{\emph{Left:} Same results as in \cref{fig:sl} (middle), but with the fraction of correct
   data points $f$ on the x-Axis.
   \emph{Right:} Multiroom example layout. The red agent needs to reach the green square, avoiding
   walls (grey) and passing through doors (blocks with coloured outline).
   }
   \label{fig:sl_f}
\end{figure}

Here we provide additional training details and results for the supervised learning experiments
performed on the CIFAR-10 dataset.
We used a ResNet18 architecture without Batchnorm, hyper-parameters for the SGD optimiser are given
in \cref{tab:sl_params}.
In \cref{tab:sl} we provide exact numerical values for the results in \cref{fig:sl-nonstat}.
We also provide values for the \emph{relative} change in error rate due to the introduction of
non-stationarities, for which the test-performance is also more affected than the train performance.

In \cref{fig:sl_f}, we show the same results as in \cref{fig:sl}, but here showing the $f$ values
used to generate $\mathcal{D}_{f,m}$ on the x-Axis.
The same `dips' in performance are visible, however from this figure it is clear that
\texttt{Dataset Size} experiences it for much smaller values of $f$, which is unsurprising, giving
the missing influences of a diverse input-data distribution.

\begin{figure}[ht]
\vspace{-0.9em}
   \centering
   \begin{subfigure}{0.32\columnwidth}
       \includegraphics[width=\linewidth]{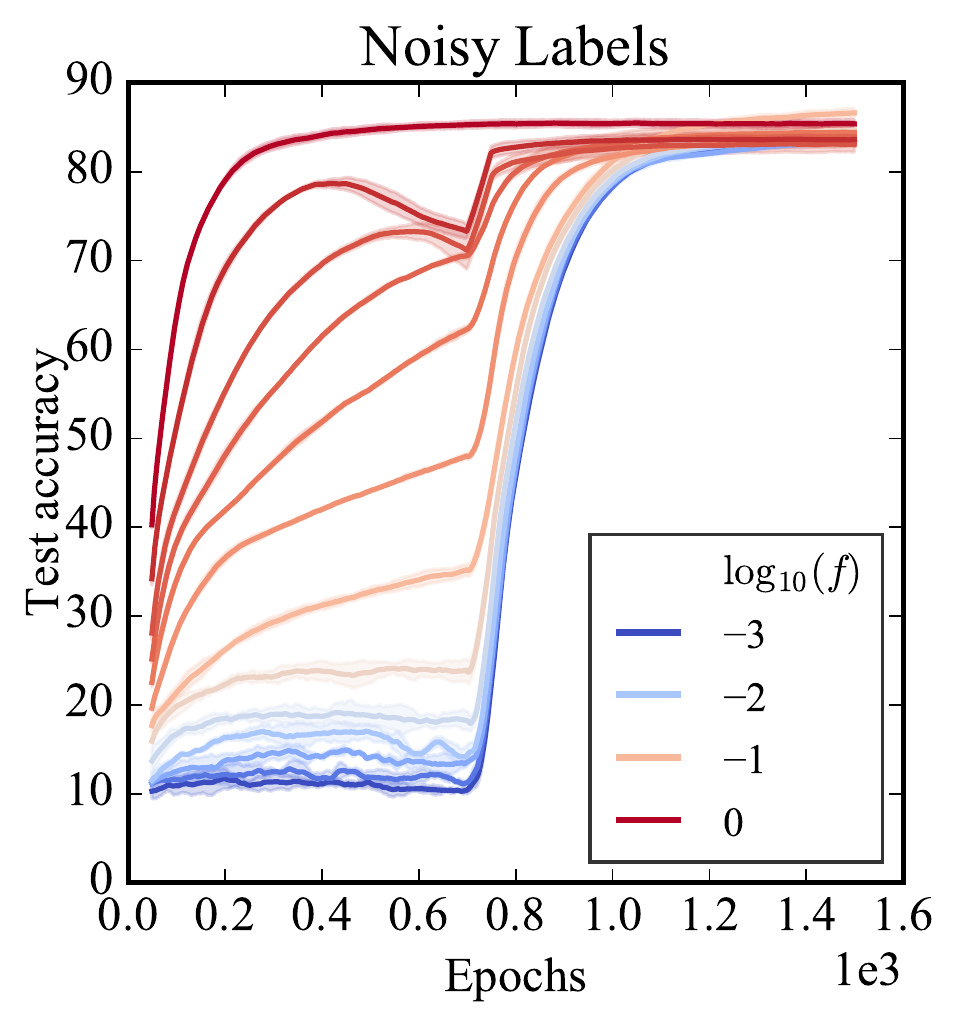}
   \end{subfigure}
   \begin{subfigure}{0.32\columnwidth}
       \includegraphics[width=\linewidth]{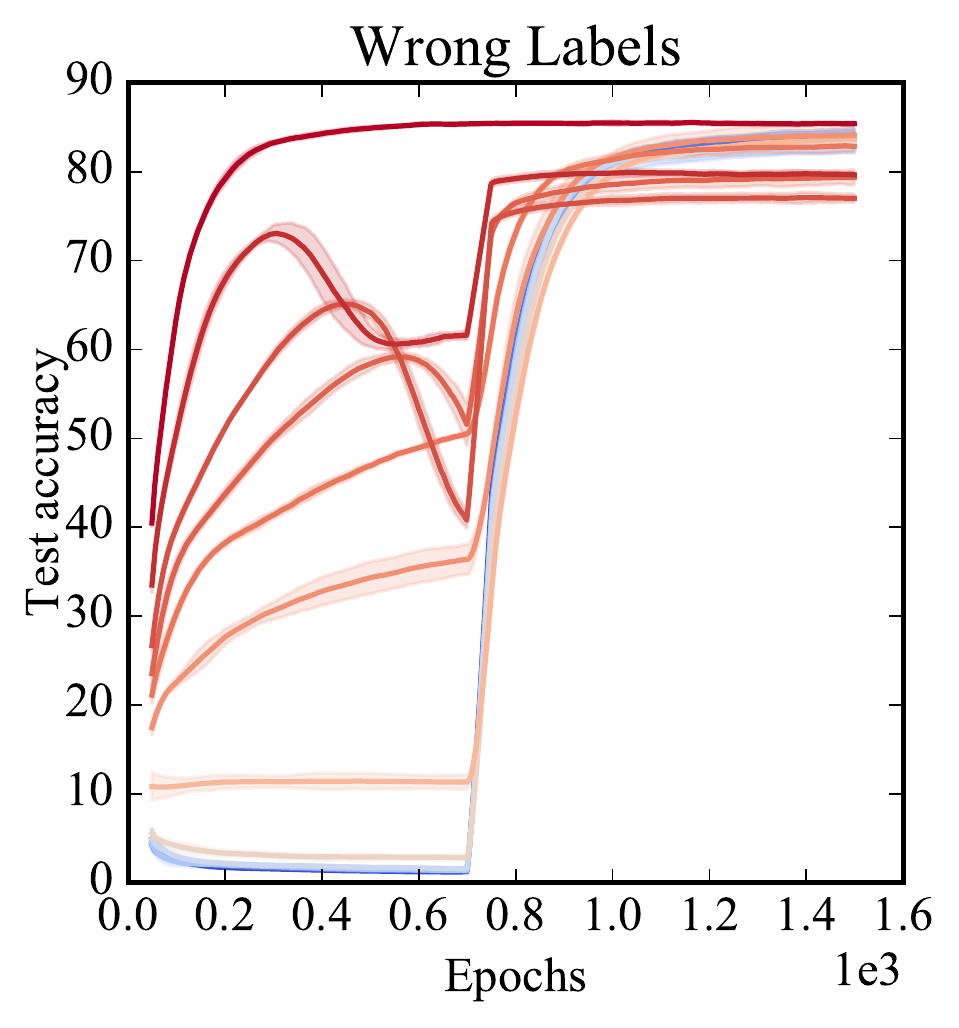}
   \end{subfigure}
   \begin{subfigure}{0.32\columnwidth}
       \includegraphics[width=\linewidth]{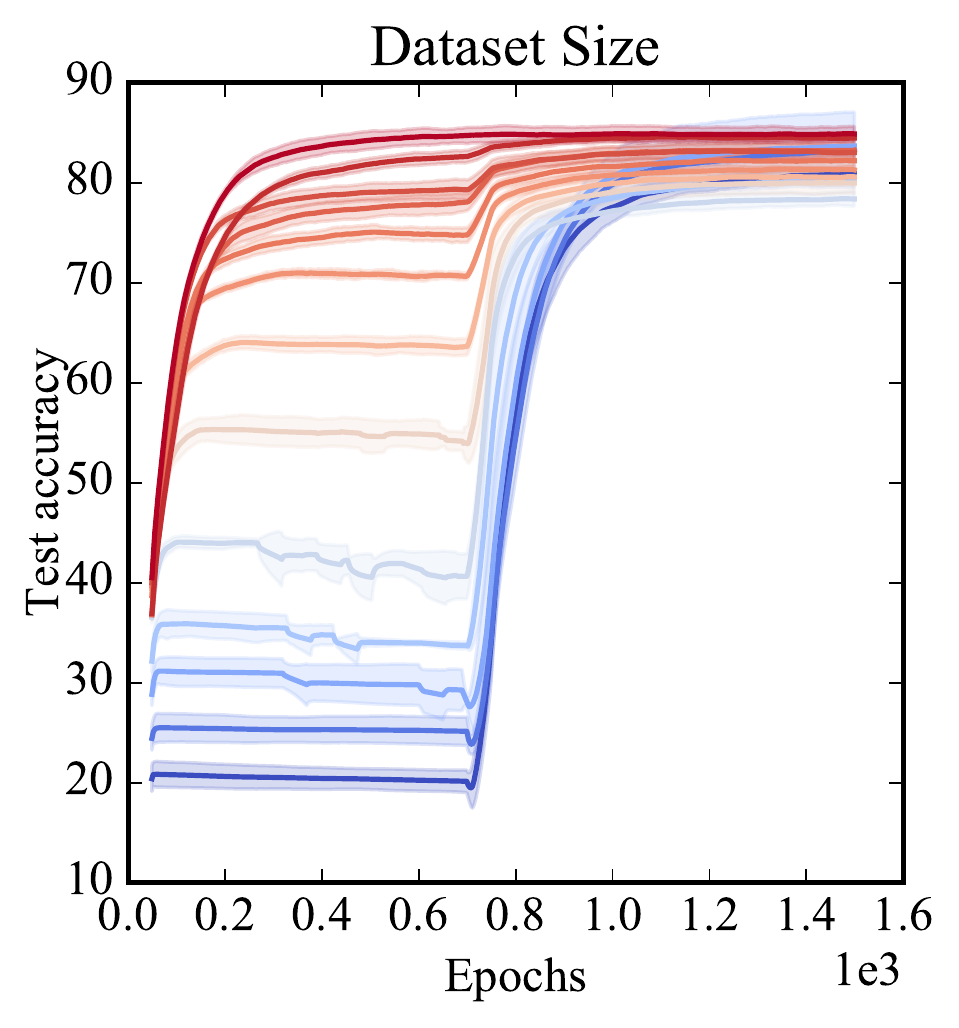}
   \end{subfigure}\\
   \begin{subfigure}{0.32\columnwidth}
       \includegraphics[width=\linewidth]{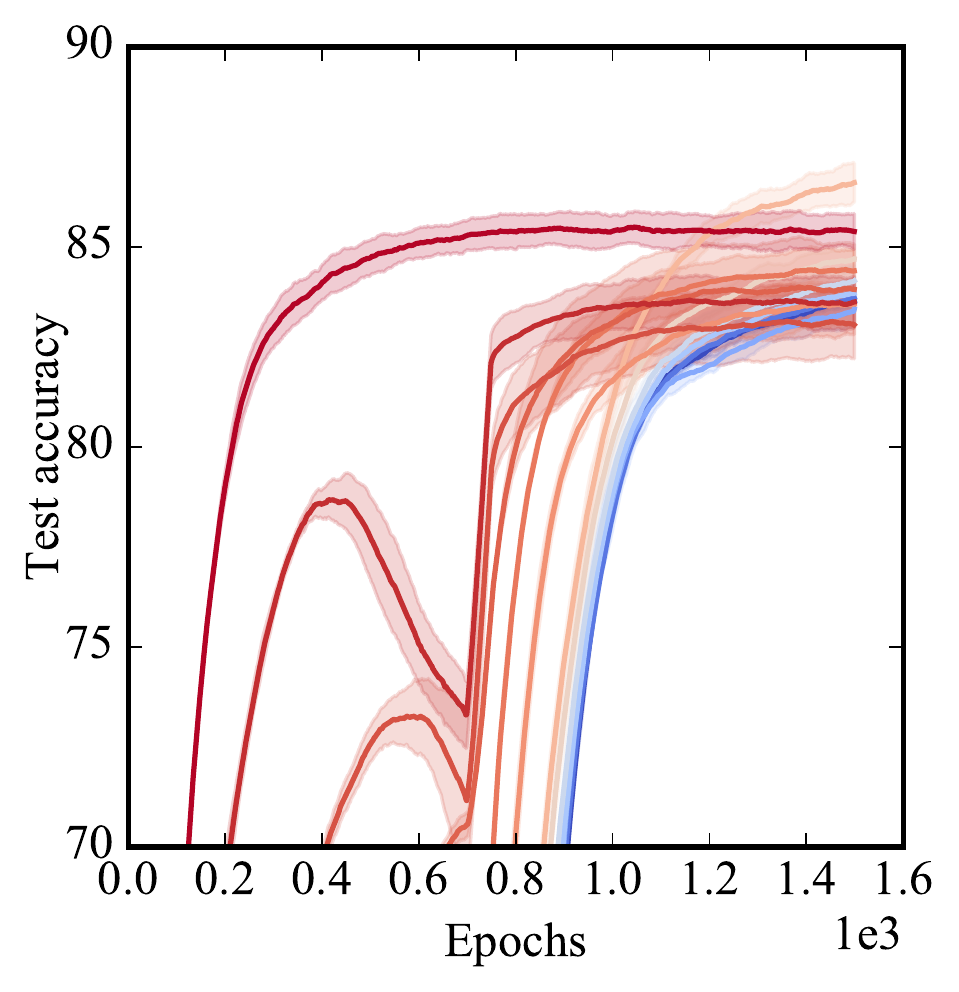}
   \end{subfigure}
   \begin{subfigure}{0.32\columnwidth}
       \includegraphics[width=\linewidth]{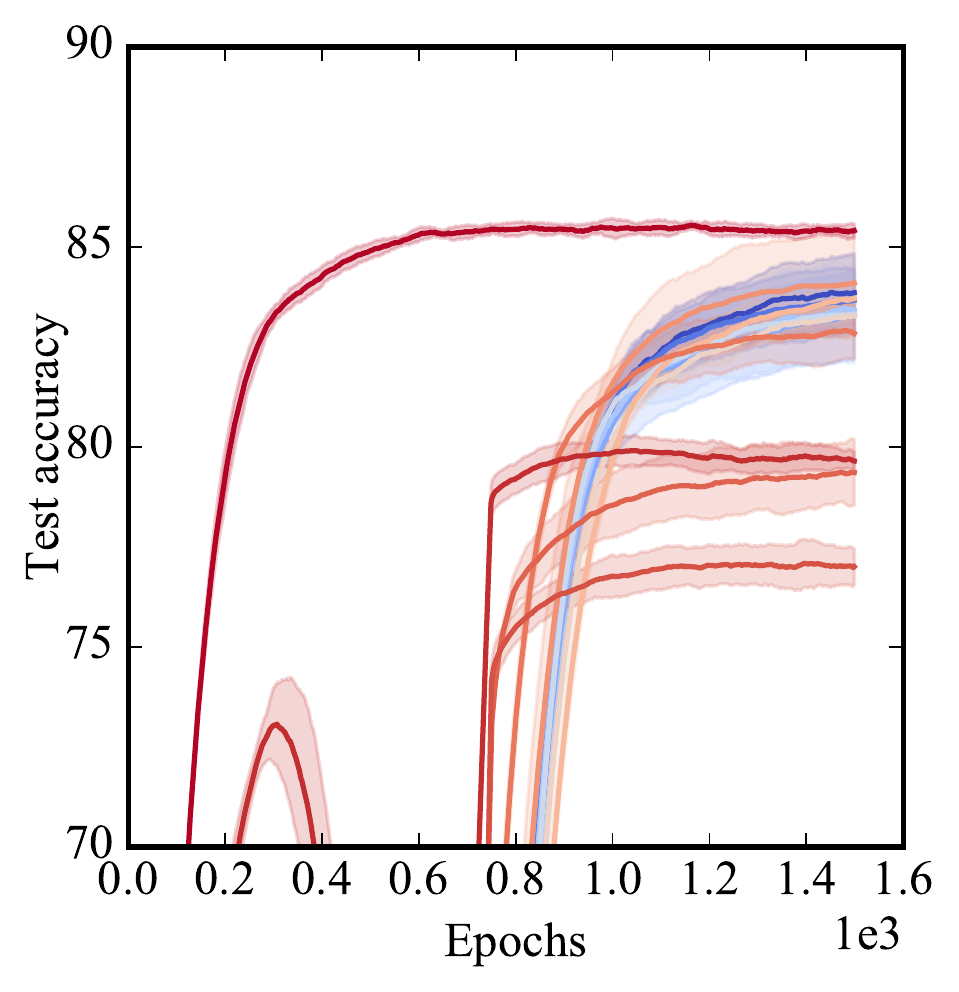}
   \end{subfigure}
   \begin{subfigure}{0.32\columnwidth}
       \includegraphics[width=\linewidth]{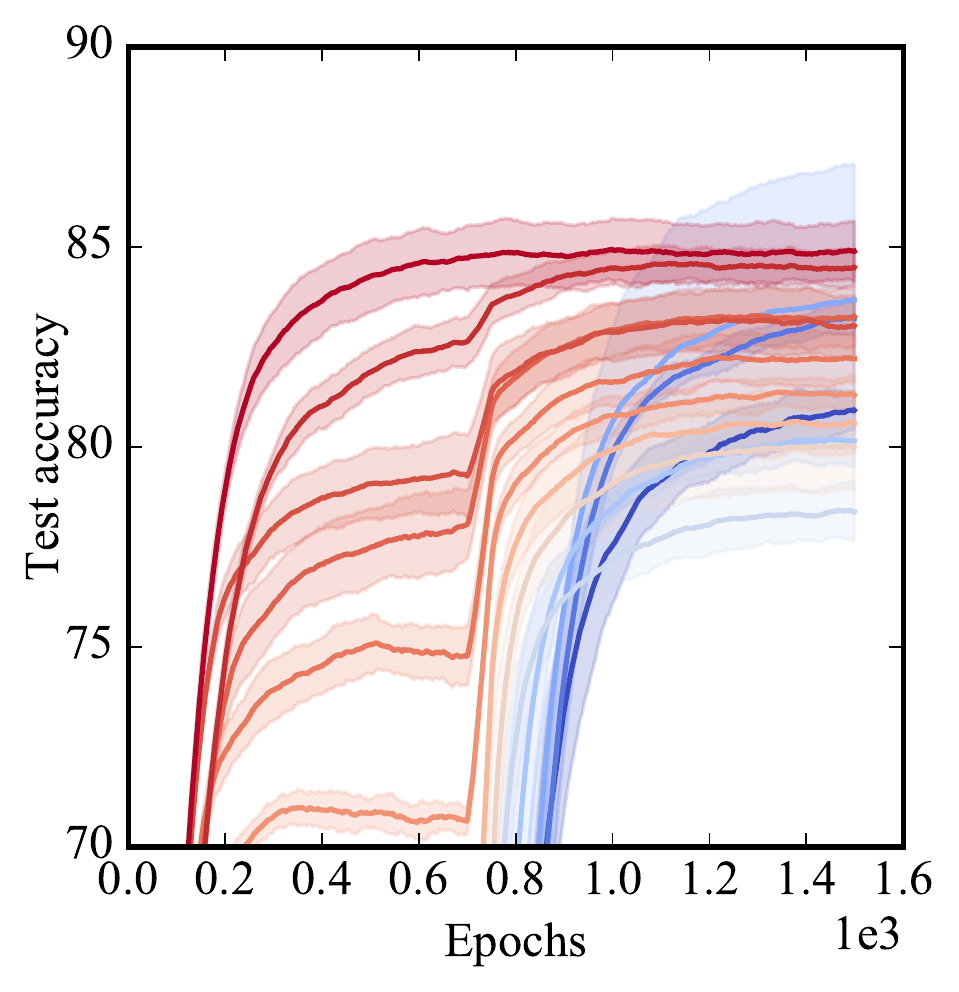}
   \end{subfigure}\\
   \caption{Individual training curves for the data used in \cref{fig:sl}(middle) and
   \cref{fig:sl_f}.
   The bottom row shows the same data as the top row, just `zoomed in'.
   }
   \label{fig:sl_detailed_training}
\end{figure}

Lastly, in \cref{fig:sl_detailed_training}, we provide the individual training runs used to generate
\cref{fig:sl}(middle) and \cref{fig:sl_f}.

\subsection{Multiroom}

In \cref{tab:multiroom} we show the hyper-parameters used for the Multiroom experiments which are
shared between `PPO' and `PPO+ITER'.
We note that our Multiroom environment uses the same modification that was used in
\citep{igl2019generalization} to make the environment fully observable.
In the original environment, the agent only observed its immediate surrounding from an ego-centric
perspective, thereby naturally generalising across various layouts.
Instead, full observability introduces the need to \emph{learn} how to generalise.
Our network consists of a three layer CNN With 16, 32 and 32 filters respectively, followed by a
fully connected layer of size 64.
One max-pooling layer is used after the first CNN layer.
We use $t_{init}=\num{4e7}$ and $t_{distill}=\num{4e7}$ (see algorithm \ref{alg:pseudo_code}) for the
duration of the initial \gls{RL} training phase and the following distillation phases.

\begin{table}[h]
\begin{floatrow}\BottomFloatBoxes
    \ttabbox
    {
    \caption{Hyper-parameters used for Multiroom} \label{tab:multiroom}
    }
    {
        \begin{tabular}{ll}
        \toprule
        Hyper-parameter                                                 & Value        \\
        \midrule
        PPO: $\lambda_{\text{Entropy  Loss}}$                               & $0.01$       \\
        PPO: $\lambda_{\text{TD}}$                                  & $0.5$        \\
        PPO: $\epsilon_{\text{Clip}}$                                   & $0.2$        \\
        PPO Epochs                                                     & 4            \\
        PPO Minibatch Size                                             & 512          \\
        Parallel Environments                                          & 32           \\
        Frames per Env per Update                                      & 128          \\
        $\lambda_{GAE}$                                                & $0.95$       \\
        $\gamma$                                                       & $0.99$       \\
        Adam: Learning rate                                            & $\num{3e-4}$ \\
        Adam: $\epsilon$                                               & $\num{1e-5}$ \\
        \bottomrule
        \end{tabular}
    }
    \ttabbox
    {
    \caption{Hyper-parameters used for ProcGen} \label{tab:ProcGen}
    }
    {
    \begin{tabular}{ll}
    \toprule
    Hyperparameter                                                 & Value \\
    \midrule
    PPO: $\lambda_{\text{Entropy  Loss}}$                               & $0.01$                        \\
    PPO: $\lambda_{\text{TD}}$                                   & $0.5$                         \\
    PPO: $\epsilon_{\text{Clip}}$                                   & $0.2$                         \\
    PPO Epochs                                                     & 3                             \\
    PPO Nr. Minibatches                                            & 8                             \\
    Parallel Environments                                          & 64                            \\
    Frames per Env per Update                                      & 256                           \\
    $\lambda_{GAE}$                                                & $0.95$                        \\
    $\gamma$                                                       & $0.999$                       \\
    Adam: Learning rate                                            & $\num{5e-4}$                  \\
    Adam: $\epsilon$                                               & $\num{1e-5}$                  \\
    Adam: Weight decay                                             & $\num{1e-4}$ \\
    \bottomrule
    \end{tabular}
    }
\end{floatrow}
\end{table}

\subsection{Boxoban}

\begin{figure}[ht]
   \centering
   \begin{subfigure}{0.30\columnwidth}
       \includegraphics[width=\linewidth]{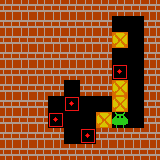}
   \end{subfigure}
   \begin{subfigure}{0.40\columnwidth}
       \includegraphics[width=\linewidth]{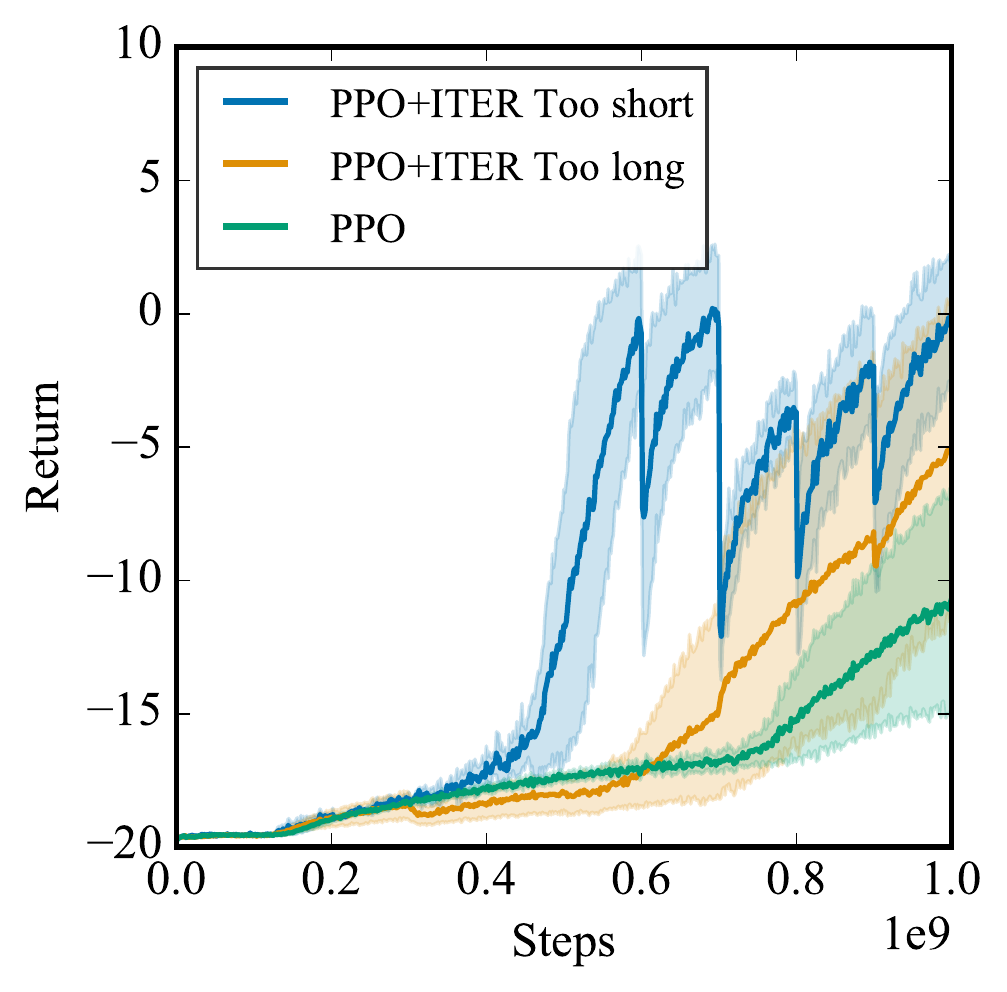}
   \end{subfigure}
   \caption{
   \emph{Left:} Boxoban example layout. The green agent needs to push (or pull) yellow boxes on the
   red targets, avoiding walls.
   \emph{Right:} Additional Boxoban results showing consequences of choosing a wrong distillation
   length, either too short or too long. Note that \texttt{ITER Too short} uses the same distillating
   length as the results in \cref{fig:results_multiroom}, but continues with distillation past
   $0.5e9$ steps.
   }
   \label{fig:ap_boxoban}
\end{figure}

For Boxoban, we re-use the same architecture and hyper-parameters as for Multiroom, but with a
reduced learning rate ($1.0e-04$) in order to stabilise training.

In \cref{fig:ap_boxoban} we show results for wrongly chosen distillation lengths. \texttt{Too short}
uses the same distillation length as the main results in \cref{fig:results_multiroom}, but continues
distilling until the end of training. Because after $0.5e9$ steps the policy performance and complexity
increase considerably, a longer distillation period would be required.
On the other hand, using a distillation period twice as long from the start (\texttt{Too long})
leads to slower training. We did not experiment with increasing the distillation length over the
course of training, since in our experiments earlier stopping of \gls{ITER} was sufficient for
optimal performance.

\subsection{ProcGen}
\label{sec:ap:ProcGen}

\begin{figure}[ht]
\vspace{-0.9em}
   \centering
   \begin{subfigure}{0.32\columnwidth}
       \includegraphics[width=\linewidth]{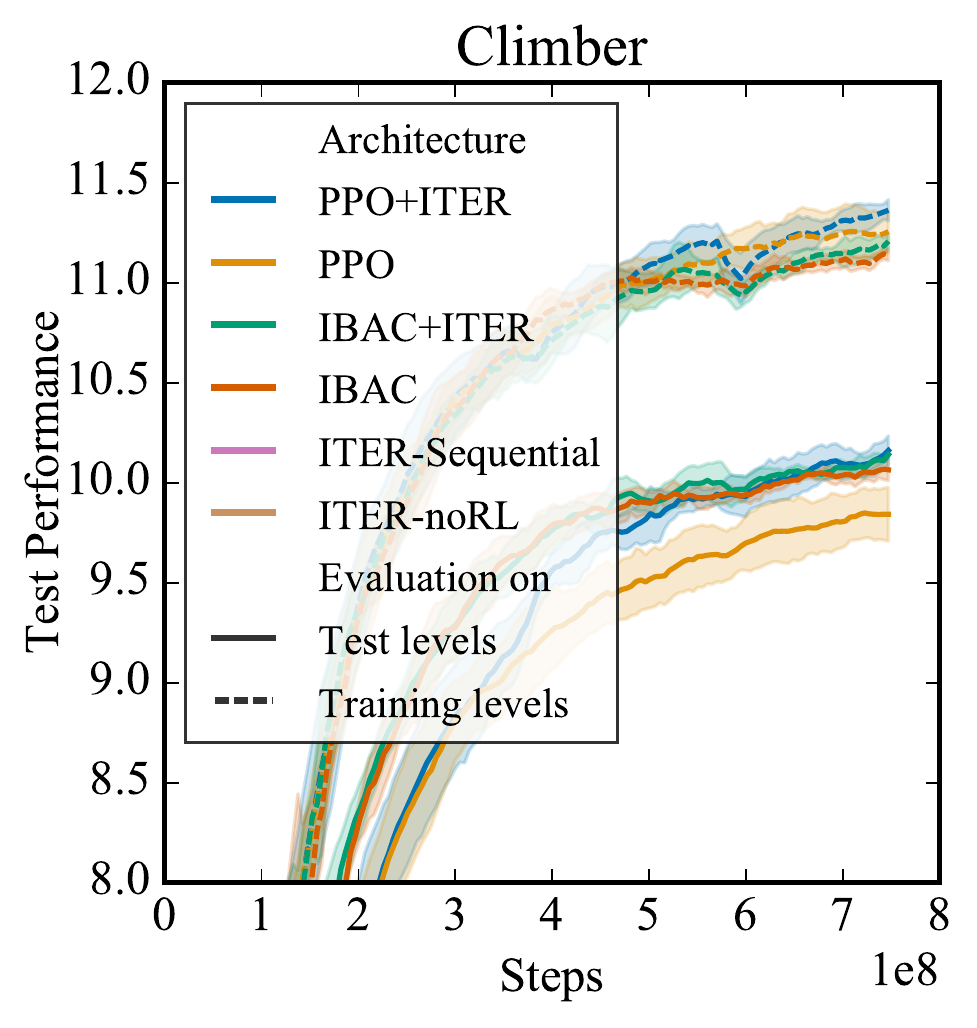}
   \end{subfigure}
   \begin{subfigure}{0.32\columnwidth}
       \includegraphics[width=\linewidth]{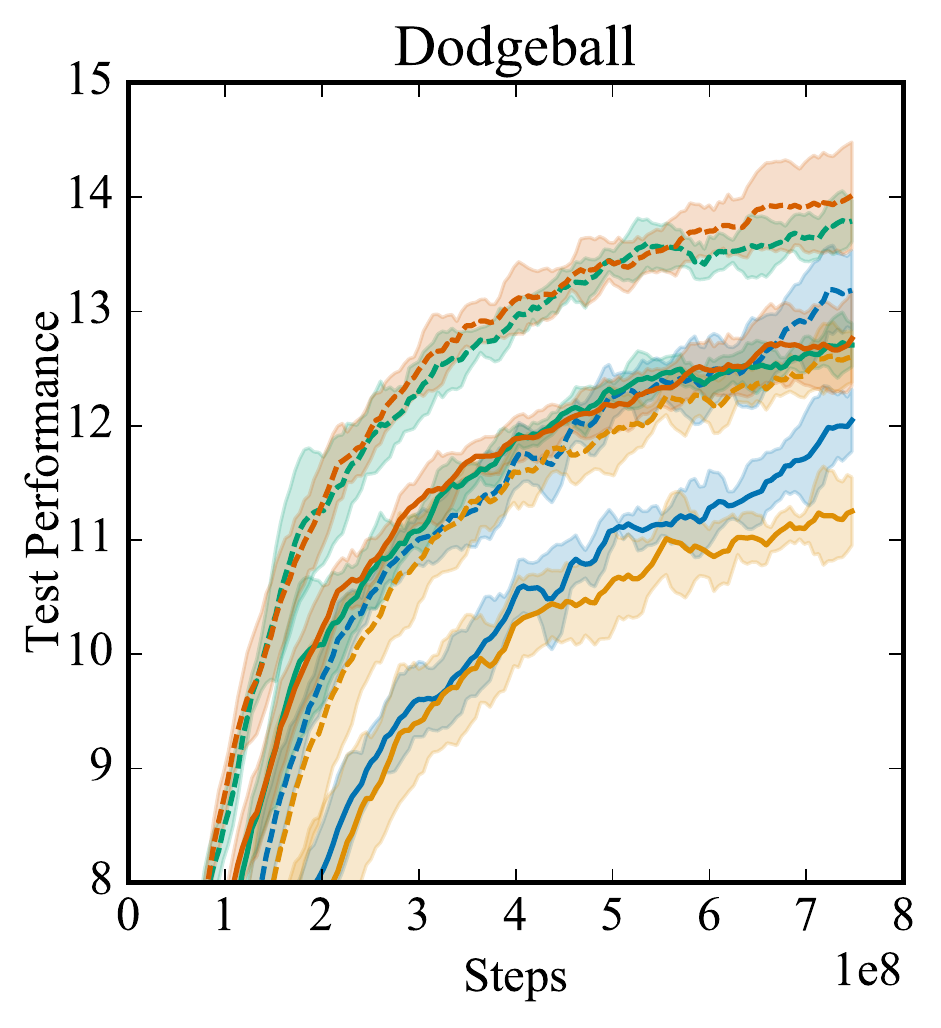}
   \end{subfigure}
   \begin{subfigure}{0.32\columnwidth}
       \includegraphics[width=\linewidth]{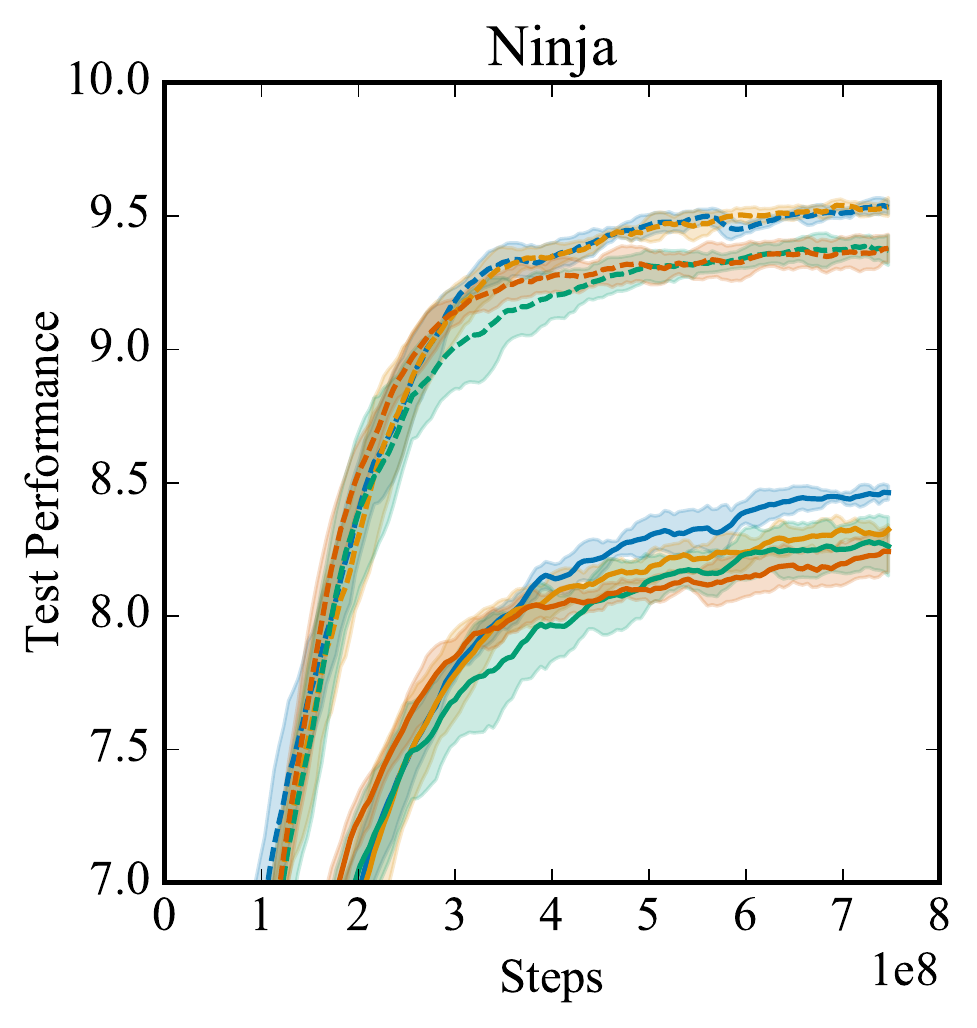}
   \end{subfigure}\\
   \begin{subfigure}{0.32\columnwidth}
       \includegraphics[width=\linewidth]{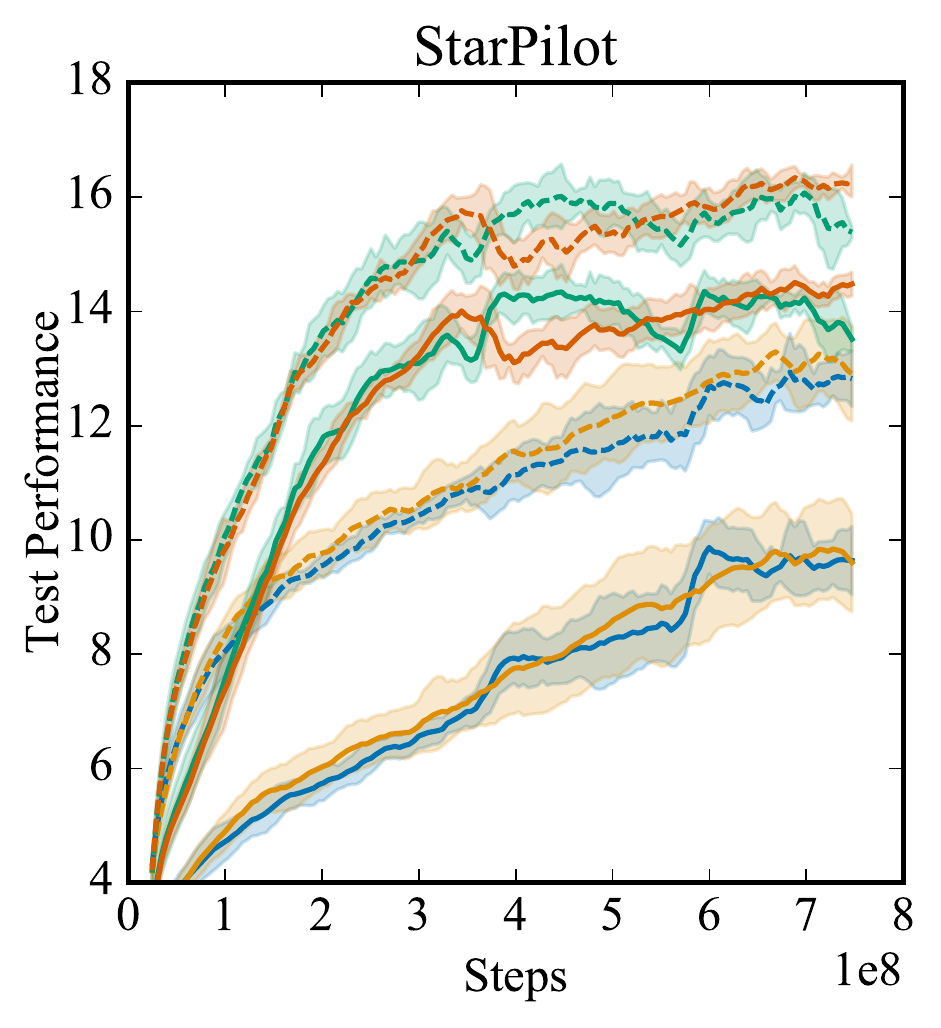}
   \end{subfigure}
   \begin{subfigure}{0.32\columnwidth}
       \includegraphics[width=\linewidth]{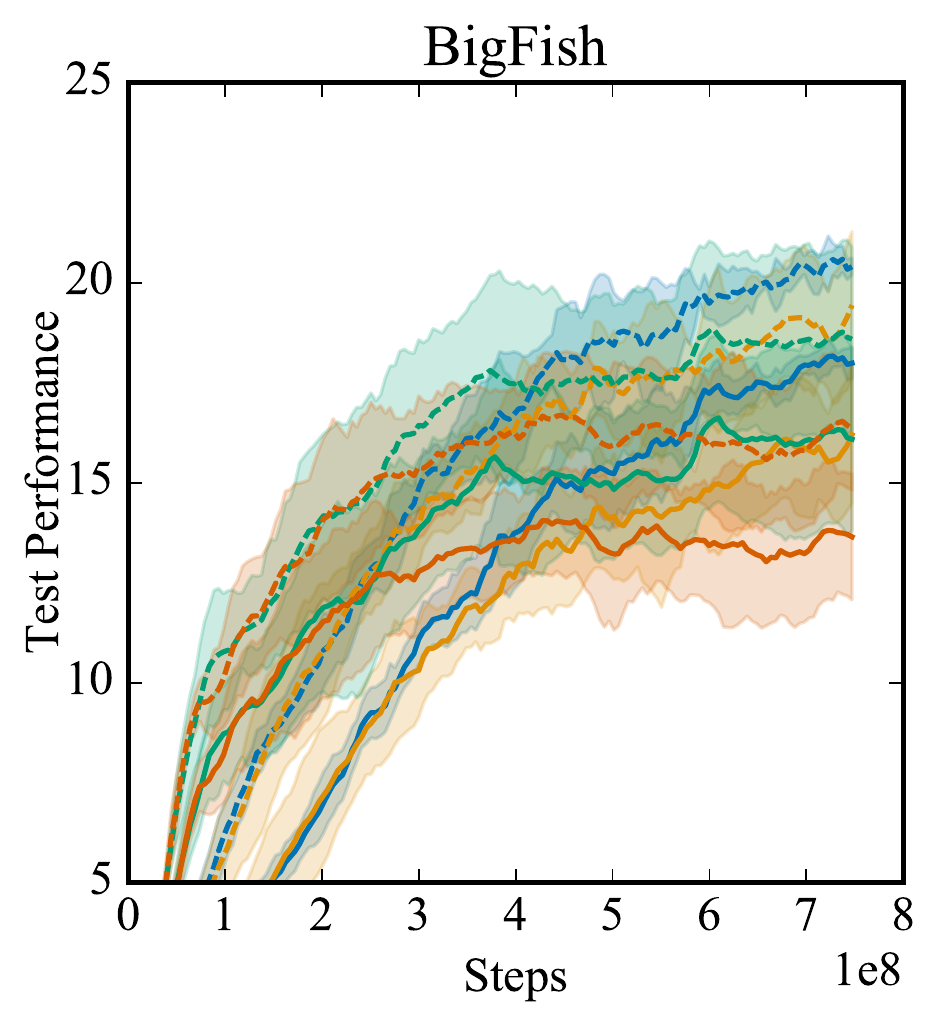}
   \end{subfigure}
   \begin{subfigure}{0.32\columnwidth}
       \includegraphics[width=\linewidth]{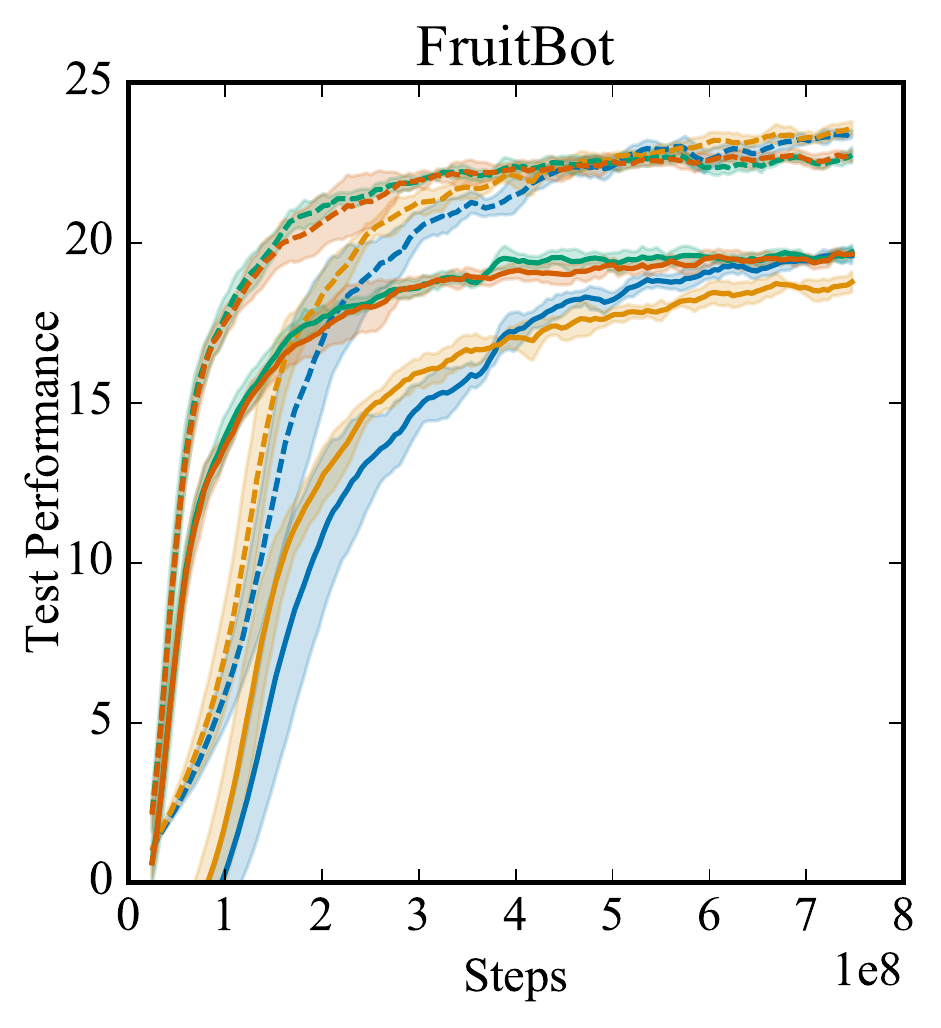}
   \end{subfigure}
   \caption{All individual results on \emph{ProcGen}. Shown is the mean and standard deviation across two random seeds.}
   \label{fig:ap:procgen}
\end{figure}

In \cref{fig:ap:procgen} we show all results on the various \emph{ProcGen} environment from which
the summary plots in the main text (\cref{fig:results_procgen,fig:ablation}) are computed.
We use the same (small) IMPALA architecture as used by \citep{cobbe2018quantifying}. Training is
done on 4 GPUs in parallel. One GPU is continuously evaluating the test performance, the other three
are used for training. Their gradients are averaged at each update step. The hyper-parameters given
in \cref{tab:ProcGen} are \emph{per GPU}. The $x$-Axis in \cref{fig:ap:procgen} shows the total number
of consumed frames, i.e. $\num{250e6}$ per \emph{training} GPU.
The distillation phase takes $t_{distill}=\num{70e6}$ frames (again per GPU) and we linearly anneal
$\alpha_\pi$ from 1 to 0 and $\alpha_V$ from 0.5 to 1.
The values of $\alpha_\pi$ and $\alpha_V$ were chosen to reflect the relative weight between $\mathcal{L}_{PG}$ and
$\mathcal{L}_{TD}$ in \cref{eq:loss} and no further tuning was done.
The initial \gls{RL} training phase takes $t_{init}=\num{50e6}$ frames.
The distillation length was chosen based on preliminary experiments on \emph{BigFish} by increasing
its length in steps of $\num{10e6}$ frames until no drop in training performance was experienced
when switching to a new student.

Due to the high computation costs of running experiments on the \emph{ProcGen} environment (4 GPUs
for about 24h for each run), we decided to exclude environments from the original benchmark based on
results presented by \citet{cobbe2019leveraging}, figures 2 and 4.
We excluded environments for two different reasons, either
because the generalisation gap was small (Chaser, Miner, Leaper, Boss Fight, Fruitbot)
or because generalisation did not improve at all during training after a very short initial jump
(CaveFlyer, Maze, Heist, Plunder, Coinrun), indicating that either it was too hard, or a very simple
policy already achieved reasonable performance.

